\documentclass{article}
\usepackage[preprint]{neurips_data_2024}

\usepackage{numprint}
\usepackage{diagbox}
\usepackage[figuresright]{rotating}
\npdecimalsign{.}
\npthousandsep{,}
\usepackage[utf8]{inputenc} 
\usepackage[T1]{fontenc}    
\usepackage{hyperref}       
\usepackage{url}            
\usepackage{booktabs}       
\usepackage{amsfonts}       
\usepackage{nicefrac}       
\usepackage{microtype}      
\usepackage{xcolor}         
\usepackage{graphicx}
\usepackage{multirow}
\usepackage{rotating}
\usepackage{adjustbox}
\usepackage{float}
\usepackage{color}

\title{A Large-scale Universal Evaluation Benchmark For Face Forgery Detection
}

\author{%
  Yijun Bei \\
  Zhejiang University\\
  \texttt{beiyj@zju.edu.cn} \\
  \And
  Hengrui Lou \\
  Zhejiang University \\
  \texttt{hengrui@zju.edu.cn} \\
  \And
  Jinsong Geng \\
  Zhejiang University \\
  \texttt{22251346@zju.edu.cn} \\
  \And
  Erteng Liu \\
  Zhejiang University \\
  \texttt{Let@zju.edu.cn} \\
  \And
  Lechao Cheng \\
  Hefei University of Technology\\
  \texttt{chenglc@hfut.edu.cn} \\
  \And
  Jie Song \\
  Zhejiang University \\
  \texttt{sjie@zju.edu.cn} \\
  \And
  Mingli Song \\
  Zhejiang University \\
  \texttt{brooksong@zju.edu.cn} \\
  \And
  Zunlei Feng \thanks{Corresponding author.}\\
  Zhejiang University \\
  \texttt{zunleifeng@zju.edu.cn} \\
}
\begin{document}
\maketitle

\begin{abstract}

With the rapid development of AI-generated content (AIGC) technology, the production of realistic fake facial images and videos that deceive human visual perception has become possible. Consequently, various face forgery detection techniques have been proposed to identify such fake facial content. However, evaluating the effectiveness and generalizability of these detection techniques remains a significant challenge. To address this, we have constructed a large-scale evaluation benchmark called DeepFaceGen, aimed at quantitatively assessing the effectiveness of face forgery detection and facilitating the iterative development of forgery detection technology.
DeepFaceGen consists of $776,990$ real face image/video samples and $773,812$ face forgery image/video samples, generated using $34$ mainstream face generation techniques. During the construction process, we carefully consider important factors such as content diversity, fairness across ethnicities, and availability of comprehensive labels, in order to ensure the versatility and convenience of DeepFaceGen. Subsequently, DeepFaceGen is employed in this study to evaluate and analyze the performance of $13$ mainstream face forgery detection techniques from various perspectives. 
Through extensive experimental analysis, we derive significant findings and propose potential directions for future research. The code and dataset for DeepFaceGen are available at \href{https://github.com/HengruiLou/DeepFaceGen} {https://github.com/HengruiLou/DeepFaceGen}.
\end{abstract}
\section{Introduction}
\label{introduciton}
In recent years, AIGC ~\cite{r1}  technology has experienced rapid development, significantly enhancing its capabilities in abstract concept learning and content generation. This technology has initiated a global wave of artificial intelligence advancements, fundamentally transforming industries such as media, entertainment, e-commerce, and education.

However, AIGC is a double-edged sword that, while revolutionizing production manner, also introduces new security risks. Malicious individuals can use AIGC to forge and tamper with data, making it difficult to verify the authenticity of generated facial images and videos ~\cite{r2}. This tampering complicates the pursuit of truth, erodes trust in multimedia information, and poses significant security threats to society. 
As a result, criminal activities such as financial scams, internet rumors, and identity theft have become increasingly widespread.

To address the misuse of deepfake facial technology, numerous researchers from both industry and academia have proposed various 
techniques for detecting face deepfakes. These techniques heavily rely on publicly available face deepfake datasets. Thus, high-quality datasets are the cornerstone for developing effective deepfake detection techniques. Recently, several deepfake datasets have been created using deepfake techniques to assist researchers in training and evaluating their detection methods. However, most current deepfake datasets focus on relatively outdated localized editing based face forgery techniques.

Recently, OpenAI released DALL·E ~\cite{dalle2020} and Sora ~\cite{sora2024}, which capable of generating full images and videos from textual descriptions, sparking a wave of full-image generation. This technology surpasses the limitations of using existing images or videos for localized edits, adopting a generative approach to creating fake content. In quick succession, numerous outstanding AIGC products have emerged, achieving unprecedented levels of generative technology. While enhancing productivity and creative efficiency, these advancements also pose significant challenges for deepfake detection research.

Therefore, some researchers adopt the diffusion based generation technology ~\cite{r3} to build the image dataset for AIGC detection. Those datasets primarily consist of general images with only a subset containing facial data, which lack significant diversity and richness in terms of facial variations. In terms of video datasets, there is a notable lack of deepfake video datasets that incorporate full-image generation based face forgery techniques, which are crucial for advancing face deepfake detection research. 
The absence of the evaluation dataset has led to a gap in face deepfake detection research, causing it to fall behind the rapid advancements in deepfake technology.

To address above challenge, this paper presents DeepFaceGen, a comprehensive and versatile evaluation benchmark specifically developed for face forgery detection. The main goal of DeepFaceGen is to facilitate the advancement of face forgery detection techniques. The benchmark encompasses a substantial dataset consisting of $463,583$ real images, $313,407$ real videos, $350,264$ forgery images, and $423,548$ forgery videos. The forgery samples are generated using 34 prevalent image/video generation techniques. Leveraging DeepFaceGen, we conduct a comprehensive evaluation of existing face forgery techniques, examining their performance across various aspects such as forgery manner, generation framework, and generalization ability. Through extensive experimentation, we uncover noteworthy insights that are anticipated to provide valuable guidance for face forgery detection tasks.

\section{Related Works}

In this section, we provide a comprehensive overview of the existing deepfake datasets, presenting detailed information summaries in Table~\ref{deepfake-dataset}. The survey of both face forgery technology and face forgery detection technology can be found in \emph{Appendix A.1}.

Early face forgery detection datasets generally suffer from a limited variety of forgery methods and are constrained in both quantity and quality. UADFV  ~\cite{r10} is the first dataset designed for face forgery detection. It only contains $49$ fake videos generated with the FakeApp application ~\cite{r20}. The construction of APFDD ~\cite{r4}, Celeb-DF ~\cite{r13}, and DeeperForensics ~\cite{r14} has significantly increased the scale of face forgery detection datasets. However, these datasets still only contain a single forgery method. To enrich the variety of forgery techniques in datasets, Korshunov \emph{et al}.~\cite{r11} developed DeepfakeTIMIT using two face swapping techniques. Subsequently, Rossler \emph{et al}. ~\cite{r12} created FF++ using a total of four forgery methods: Deepfake ~\cite{r22}, Face2face ~\cite{r23}, Faceswap ~\cite{r24}, and NeuralTextures ~\cite{r25}. However, the size and diversity of FF++ are still insufficient, making it challenging to optimally train high-performance deep models with a large number of parameters. Zi \emph{et al}. ~\cite{r15} collected deepfake samples from the internet to create WildDeepfake, which includes facial motion sequences extracted from videos. After manually removing videos without corresponding real faces, the number of fake videos stands at $3,509$. Although the visual effects are closer to real-life scenarios, the limited data volume poses constraints on training high-performance deep models.

To address the issues of poor generation quality and coarse tampering traces in early face forgery detection datasets, DFDC ~\cite{r16}, initially released as part of Facebook's eponymous competition, contains $5,250$ videos, which was later supplemented to reach $104,500$ fake videos generated using eight different methods to ensure dataset diversity. Kwon \emph{et al}. ~\cite{r17} generated the KoDF dataset, comprising $175,776$ fake videos using $6$ forgery methods. Most videos in KoDF feature Korean individuals, representing the first effort to address the underrepresentation of Asian populations in existing forgery detection datasets. FakeAVCeleb ~\cite{r18} consists of $19,500$ fake videos generated using $3$ face forgery methods: Faceswap, DeepFaceLab ~\cite{r26}, and FSGAN ~\cite{r27}. He et al. ~\cite{r19} developed ForgeryNet, the first face forgery detection dataset that includes videos and images. They used $15$ deepfake methods to generate $121,617$ fake videos and $1,457,861$ fake images. Although the forgery quantity and forgery method have been significantly improved, ForgeryNet's forgery method is still limited to the face forgery technology based on localized editing, which cannot provide data support for detecting the novel AIGC forgery method based on full-image generation. 
Additionally, ForgeryNet fails to take into account the aspects of racial fairness and fine-grained annotations, which are essential for conducting a comprehensive and multi-perspective evaluation.

The rapid development of full-image generation based face forgery techniques, exemplified by diffusion models, has led to the emergence of outstanding AIGC products such as Sora ~\cite{sora2024}, Midjourney ~\cite{midjourney2022} and DALL·E ~\cite{dalle2020}. These products have significantly impacted the field with their astonishing realism. The construction of deepfake datasets based on full-image generation techniques has become increasingly urgent due to the astonishing realism of these AIGC products. IEEE VIP Cup ~\cite{r6} and DE-FAKE ~\cite{r7} are among the first to contain diffusion models for generating images, pioneering efforts in this area. Additionally, IEEE VIP Cup includes localized editing based forgery techniques in its dataset construction, marking the first effort to build a general-purpose dataset in the field of image forgery datasets. Based on the CIFAR-10 dataset ~\cite{2009Learning}, J. Bird \emph{et al}. ~\cite{r8} created the CiFAKE dataset by generating fake images using SDV1.4. GenImage ~\cite{r9} includes 7 diffusion-based full-image generation forgeries and the localized editing forgery technology BigGAN ~\cite{brock2018large}, covering $1,350,000$ images across $1,000$ categories. In the context of the proliferation of full-image generation forgeries, these datasets provide valuable image evaluation data for researchers. However, the quantity and quality of facial data in these datasets are difficult to guarantee. They do not provide dedicated data for deepfake detection research on facial images and lack datasets that include full-image generated facial forgeries within face video forgery datasets.

\begin{table}[!t]
    \centering
    \caption{
Summary of existing deepfake datasets.
    }
    \label{deepfake-dataset}
    \resizebox{\textwidth}{!}{
    \begin{tabular}{lccccccccc}
        \toprule
         \multirow{2}{*}{Dataset Name} & \multirow{2}{*}{Content} & \multicolumn{2}{c}{Forged Data} & \multicolumn{2}{c}{Generation Manner} &  Racial & Fine-grained &Forgery &Public\\
         \cmidrule(lr){3-4}
         \cmidrule(lr){5-6}
        & & Image& Video & Localized & Full-image &Balance & Annotation&Approaches & Avalibility \\
                \midrule
        APFDD~\cite{r4} & Face & 5,000 & - & $\checkmark$ & $\times$ & $\times$ & $\times$ & 1 & $\times$ \\
        DeepArt~\cite{r5} & Art & 73,411 & - & $\times$ & $\checkmark$ & $\times$ & $\times$ & 5 & $\checkmark$ \\
        IEEE VIP Cup~\cite{r6} & General & 7,000 & - & $\checkmark$ & $\checkmark$& $\times$ & $\times$ & 14 & $\times$ \\
        DE-FAKE~\cite{r7} & General & 60,000 & - & $\times$ & $\checkmark$ & $\times$ & $\times$ & 4 & $\times$ \\
        CiFAKE~\cite{r8} & General & 60,000 & - & $\times$ & $\checkmark$ & $\times$ & $\times$ & 1 & $\checkmark$ \\
        GenImage~\cite{r9} & General & 1,350,000 & - & $\checkmark$ & $\checkmark$ & $\times$ & $\times$  & 8 & $\checkmark$\\
        \midrule
        UADFV~\cite{r10} & Face & - & 49 & $\checkmark$ & $\times$ & $\times$ & $\times$ & 1 & $\times$ \\
        DeepfakeTIMT~\cite{r11} & Face & - & 320 & $\checkmark$ & $\times$ & $\times$ & $\times$ & 2 & $\checkmark$ \\
        FF++~\cite{r12} & Face & - & 4,000 & $\checkmark$ & $\times$ & $\times$ & $\checkmark$ & 4 & $\checkmark$  \\
        Celeb-DF~\cite{r13} & Face & - & 5,639 & $\checkmark$ & $\times$ & $\times$ & $\checkmark$  & 1 & $\checkmark$  \\
        DeeperForensics~\cite{r14} & Face & - & 10,000 & $\checkmark$ & $\times$ & $\times$ & $\times$ & 1 & $\checkmark$ \\
        WildDeepfake~\cite{r15} & Face & - & 3,509 & $\checkmark$ & $\times$ & $\times$ & $\times$ & unknown & $\checkmark$ \\
        DFDC~\cite{r16} & Face & - & 104,500 & $\checkmark$ & $\times$ & $\times$ & $\times$ & 8 & $\checkmark$ \\
        KoDF~\cite{r17} & Face & - & 175,776 & $\checkmark$ & $\times$ & $\times$ & $\times$ & 6 &$\times$\\
        FakeAVCeleb~\cite{r18} & Face & - & 19,500 & $\checkmark$ & $\times$ & $\times$ & $\times$ & 3 & $\checkmark$ \\
        \midrule
        ForgeryNet~\cite{r19} & Face & 1,457,861 & 121,617 & $\checkmark$ & $\times$ & $\times$ &$\times$  & 15 & $\checkmark$ \\
         \textbf{DeepFaceGen (ours)} &  \textbf{Face} &  \textbf{350,264} &  \textbf{423,548} &  \textbf{$\checkmark$} &  \textbf{$\checkmark$} &  \textbf{$\checkmark$} &  \textbf{$\checkmark$} &  \textbf{34} &  \textbf{$\checkmark$} \\
         \bottomrule
    \end{tabular}}
\end{table}

\section{Evaluation Dataset Construction}

In this section, we aim to construct a robust and extensive benchmark for the detection of face forgery. To accomplish this, we carefully consider a range of critical factors including the manner of generation, generation framework, content diversity, ethnic fairness, and label richness throughout the benchmark development process. Following this, we provide a detailed introduction to the methodologies employed for collecting and generating forged samples. Additionally, we introduce the authentic data sources utilized by DeepFaceGen. 
Lastly, we present a comprehensive summary of the detailed data information encompassed within DeepFaceGen.

To enhance the diversity of DeepFaceGen, we augment its dataset by incorporating a selection of pre-existing forged face samples alongside newly generated ones using popular image and video generation techniques. These collected samples adhere to the principle of ethnic fairness. Specifically, from references ~\cite{r13} and ~\cite{r19}, we choose samples created through localized editing techniques such as face swapping, face reenactment, and face alteration. Detailed information about these collected samples can be found in \emph{Appendix A.2}.

\subsection{Forged Face Sample Generation}

For the novel AIGC techniques, we employ a set of $17$ prevalent full-image generation based face forgery techniques. Additionally, we incorporate $17$ classical localized editing based face forgery techniques, excluding the new generation methods. In the following section, we extensively elaborate on the generation processes for both categories of techniques.

\textbf{Full-image Based Generation} techniques utilize text or image input to generate full-image samples. The design of the prompt plays a crucial role in determining the quality of the generation outcome. Hence, we primarily present the process of prompt construction, followed by the description of forgery methods.

\begin{itemize}
  \item \textbf{Prompts Construction}. 
  In the design of prompts, we strive to achieve both content diversity and fairness, which are accompanied by a strong emphasis on detailed prompt descriptions. For each prompt, we establish fundamental attributes, such as age, gender, and skin tone, while also providing comprehensive specifications regarding the person's background and physical features. The inclusion of these extensive textual attribute details further facilitates the evaluation of forgery detection performance at a fine-grained level. A total of $9$ textual attributes are defined in the prompt construction process. By exhaustively generating prompts using all possible combinations of these textual attributes, we ensure the creation of a diverse and equitable set of forged data. For further elaboration on these prompts, please refer to \emph{Appendix A.3}.
  \item  \textbf{Text2Image} generation techniques involve three main categories: GAN, autoregressive, and diffusion frameworks. Some of these techniques have been developed into commercial products. In order to enhance the practicality and universality of DeepFaceGen, we have incorporated mature commercial products and popular open-source methods to generate the forgery samples.
    For GAN-based models, we have adopted the popular open-source DF-GAN ~\cite{DFGAN} which employs adversarial training between the generator and discriminator to achieve impressive image generation capabilities. As for autoregressive based models, we have utilized OpenAI's commercial product DALL·E and DALL·E 3~\cite{dalle2020}, which treats text tokens and image tokens as a unified data sequence and uses a Transformer for auto-regression.
    Given that existing high-quality generation techniques mostly rely on diffusion framework, we have incorporated specific models such as OpenAI's Midjourney~\cite{midjourney2022}, Baidu's Wenxin ~\cite{wenxin2022}, Stability.ai's series products \{Stable Diffusion 1 (SD1), Stable Diffusion 2 (SD2), Stable Diffusion XL (SDXL)\} ~\cite{SD}, and PromptHero's open-source version of Midjourney (Openjourney, OJ) ~\cite{openjourney}. To ensure that these models are suitable for face generation, we fine-tune pre-trained open-source text2video models using LoRA~\cite{hu2022lora}.

  \item 
  \textbf{Image2Image} generation involves utilizing an image as input to generate full-image samples, typically employing diffusion frameworks. Therefore, we utilize Stable Diffusion XL Refiner (SDXLR), Stable Diffusion InstructPix2Pix (Pix2Pix), and Stable Diffusion ImageVariation (VD) ~\cite{SD}, all of which have achieved high rankings on Huggingface's download charts.
  
  \item 
  \textbf{Text2Video} techniques involve using a text prompt as input to generate a complete video sample, also relying on diffusion frameworks. However, due to unavailability of certain mature commercial products' API, we have selected alternative products. Specifically, we have chosen MagicTime ~\cite{yuan2024magictime},  AnimateDiff-Lightning (AnimateDiff) ~\cite{ACMlightning}, AnimateLCM ~\cite{wang2024animatelcm}, Hotshot ~\cite{Mullan_Hotshot-XL_2023}, and Zeroscope ~\cite{zeroscope_v2_576w}.

\end{itemize}

\textbf{Localized Editing Based Generation} technique generates forged samples by modifying certain parts of input face images. Existing Localized Editing techniques can be categorized into three types: face swapping, face reenactment, and face alteration.
\begin{itemize}
  \item \textbf{Face Swapping} technique involves creating a manipulated face sample by exchanging the faces of two given image samples. In this study, we employ $8$ commonly used face swapping methods, namely FaceShifter ~\cite{Li2019FaceShifterTH}, FSGAN ~\cite{r27}, DeepFake ~\cite{r22}, BlendFace ~\cite{blendface}, MMReplacement ~\cite{r19}, DeepFakes-StarGAN-Stack (DSS), StarGAN-BlendFace-Stack (SBS), and SimSwap ~\cite{Chen2020SimSwapAE}. Among these approaches, DSS and SBS are categorized as mixed face forgery methods, wherein the face alteration technique is initially applied before face swapping is performed.
  \item  \textbf{Face Reenactment} technique involves transferring the facial movements and expressions from one person onto the face of another person. In this study, we utilize four specific approaches for face reenactment: Talking Head Video ~\cite{talkinghead}, ATVG-Net ~\cite{atvgnet}, FOMM ~\cite{FOMM}, and Motion-cos ~\cite{motion}.
  \item \textbf{Face Alteration} technique involves creating forged images by making subtle modifications to facial attributes such as hair color, beard, and glasses. The face alteration approaches utilized in this study include StyleGAN2 ~\cite{StyleGAN2}, MaskGAN ~\cite{maskgan}, StarGAN2 ~\cite{StarGAN2}, SC-FEGAN ~\cite{scfegan}, and DiscoFaceGAN ~\cite{DiscoFaceGAN}.
\end{itemize}

\subsection{Authentic Face sample Collection}
\label{dataset_source}

In order to ensure content diversity and ethnic fairness in the authentic face samples used in DeepFaceGen, we obtained real samples from reputable sources including ~\cite{r13}, ~\cite{r19}, ~\cite{CN-CVS}, and ~\cite{CMLR}. The final collection consists of $463,583$ images and $313,407$ videos, encompassing diverse races, genders, ages, expressions, hairs, backgrounds, and so on. 
Please refer to \emph{Appendix A.2} for more details.

\subsection{Dataset Summarization}

The aforementioned generation and collection processes yield the initial dataset samples. To ensure both sample quality and racial balance, postprocess operations are implemented to filter these samples. The SkinToneClassifier ~\cite{skin} is utilized for racial balance, ensuring skin tone balance in the generation and collection of localized editing-based and Image2Image face forgery methods. For full-image generation-based face forgery techniques (Text2Image and Text2Video), the combination and design of text prompts also take skin tone balance into consideration. Additionally, we employ YOLO ~\cite{yolo} with manual screening to eliminate low-quality data. These measures effectively maintain the fairness and reliability of DeepFaceGen, resulting in the collection of $350,264$ forged images and $423,548$ forged videos. For a detailed breakdown of the sample numbers for different generation techniques, please refer to Figure 2 provided in \emph{Appendix A.2}.

\section{Benchmark Evaluation and Analysis}

In this section, we employ DeepFaceGen to evaluate $13$ prevalent face forgery detection methods from various perspectives, such as generation approach type, generalization capability, and technique relevance. Subsequently, we analyze extensive experimental results and summarize key findings, elucidating the strengths and weaknesses of current face forgery detection techniques, as well as identifying potential directions for future research.

\textbf{Evaluation Settings.}
\label{settings}
Based on the distinction in modality between images and videos, we partition DeepFaceGen into two parts. The image and video datasets are divided into training, validation, and test subsets in a ratio approximately $7:1:2$. To ensure fairness in evaluation, each subset maintains a ratio of real to fake instances close to $1:1$. For image-level assessments, we employ Xception ~\cite{Xception}, EfficientNet-B0 ~\cite{tan2020efficientnet}, F3-Net ~\cite{F3NET}, RECCE ~\cite{recce}, DNANet ~\cite{dnaet}, and FreqNet ~\cite{tan2024frequencyaware}. For video-level evaluations, we select MesoNet ~\cite{Afchar2018MesoNetAC}, EfficientNet-B0 ~\cite{tan2020efficientnet}, Xception ~\cite{Xception}, F3-Net ~\cite{F3NET}, CViT ~\cite{cvit}, SLADD ~\cite{sladd}, and Exposing ~\cite{ba2024exposing}, as they exhibit exceptional performance in forgery video detection. The experiments are conducted separately on a machine equipped with an Nvidia A40 GPU (48GB VRAM) and two machines, each featuring a GeForce RTX 4090 GPU (24GB VRAM). More evaluation details are given in the \emph{Appendix A.4}.

\subsection{Evaluation of Mainstream Forgery Detection Techniques}
\label{sec:capability-assessment}
In this section, we initiate the training of all forgery detection models utilizing training samples obtained from DeepFaceGen. We subsequently present and analyze the experimental results comprehensively, considering various aspects such as the sample modality, forgery technique, forgery technique type, and the framework employed by the forgery detection models.

\subsubsection{Image-level Evaluation and Analysis}
\begin{figure}[!t]
  \centering
  \includegraphics[width=\linewidth]{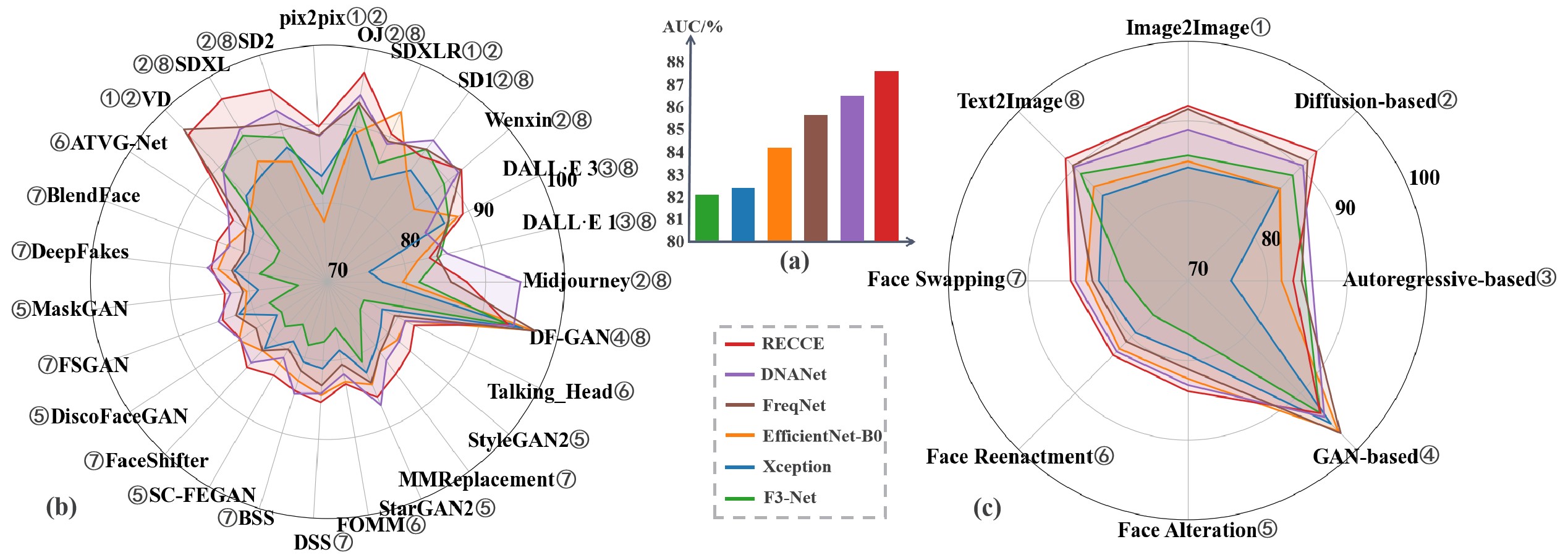} 
  \caption{Image-level Performance comparison of different forgery detection techniques. (a) Average detection performance ranking. (b) Detection performance for different generation techniques. (c) Detection performance for different generation manners and frameworks (marked with \textcircled{1}-\textcircled{8}). 
   }
  \label{fig5}
\end{figure}

\textbf{Forgery Detection Technique Comparison.} 
Figure~\ref{fig5} (a) illustrates the average detection performance of various forgery detection techniques. It is evident from the figure that RECCE \cite{recce} and DNANet \cite{dnaet} outperform the rest, while Xception \cite{Xception} and F3-Net \cite{F3NET} exhibit poor performance. RECCE utilizes an Encoder-Decoder structure for forgery detection, while DNANet \cite{dnaet} combines an encoder classifier and a contrastive learning projector. These two methods leverage their respective architectures to extract detailed features related to the forgery. FreqNet \cite{tan2024frequencyaware}, on the other hand, extracts high-frequency features across both spatial and channel dimensions, capturing subtle forgery variations. In contrast, general-purpose classifiers like Xception \cite{Xception}, F3-Net \cite{F3NET}, and EfficientNet-B0 \cite{tan2020efficientnet}, which employ convolutional encoder architectures, do not perform as well as the specialized methods designed for face forgery detection. Therefore, it can be inferred that \emph{ the details extraction module plays a critical role in the detection of face image forgery (\textbf{Finding 1})}.

\textbf{Generation Manner and Framework.} 
Based on Figure~\ref{fig5} (c), it is evident that localized editing techniques (face swapping, face reenactment, face alteration) for image generation can produce more challenging identification samples compared to full-image generation techniques (Text2Image and Image2Image). This can be attributed to \emph{the relative ease of generating authentic images by modifying smaller localized areas rather than the entire image (\textbf{Finding 2})}. However, further research is required to enhance the performance of full-image generation techniques.
Regarding different generation frameworks, it is evident that autoregressive based techniques (DALL·E and DALL·E3 ~\cite{dalle2020}) achieve the highest quality of forgery, surpassing diffusion-based and GAN-based techniques. The newly proposed diffusion-based framework demonstrates the second-best average performance, indicating its potential for further development. Conversely, GAN-based generation techniques exhibit the poorest quality for forgery. Therefore, it can be concluded that \emph{autoregressive-based and diffusion-based generation techniques are capable of producing more realistic forged face images than GAN-based generation techniques (\textbf{Finding 3})}.

\textbf{Input Modality.} 
Based on the results depicted in Figure~\ref{fig5} (b)(c), it is apparent that both Text2Image (Midjourney~\cite{midjourney2022}, OJ~\cite{openjourney}, SD1~\cite{SD}, SD2~\cite{SD} and SDXL~\cite{SD}) and Image2Image techniques (SDXLR~\cite{SD}, Pix2Pix~\cite{SD}, and VD~\cite{SD}) that employ the same diffusion-based framework deliver comparable performance. Consequently, it can be deduced that \emph{the choice of input modality has minimal influence on the quality of image generation (\textbf{Finding 4})}.

\begin{figure}[!t]
  \centering
  \includegraphics[width=\linewidth]{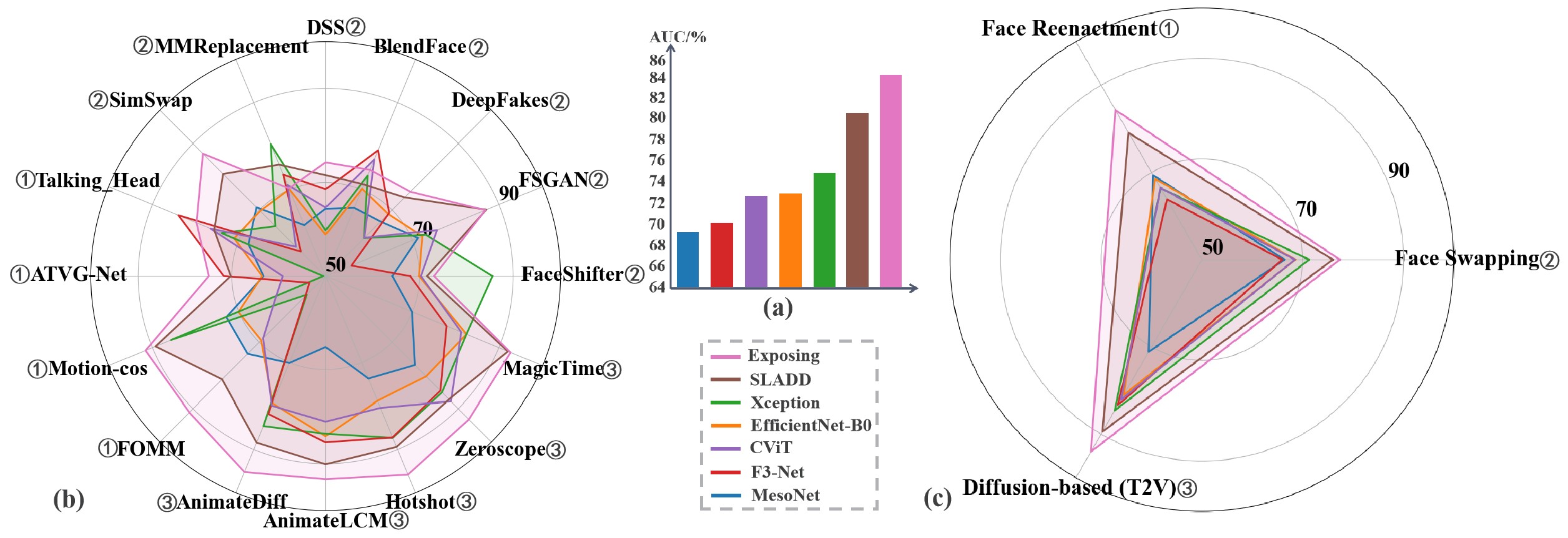} 
  \caption{
  Video-level Performance comparison of different forgery detection techniques. (a) Average detection performance histogram. (b) Detection performance for different generation techniques. (c) Detection performance for different generation manners and frameworks (marked with \textcircled{1}-\textcircled{3}).
  }
  \label{fig66}
\end{figure}

\subsubsection{Video-level Evaluation and Analysis}

\textbf{Forgery Detection Technique Comparison.} 
Figure~\ref{fig66} (a) depicts the average detection performance of various video forgery detection techniques. It is evident that both Exposing~\cite{ba2024exposing} and SLADD~\cite{sladd} outperform the other techniques. Exposing~\cite{ba2024exposing} adopts a two-step approach: extracting frame-level facial bounding boxes from raw videos and subsequently extracting multiple disentangled local features from different regions for forgery detection. SLADD~\cite{sladd} employs adversarial self-supervised training to identify various forgery detail features, which contributes to its outstanding performance. In contrast, general-purpose classifiers such as EfficientNet-B0~\cite{tan2020efficientnet}, Xception \cite{Xception}, F3-Net \cite{F3NET}, and CViT~\cite{cvit} exhibit poor identification performance due to their lack of forgery detail information.
Thus, we can conclude that the \emph{extraction of detailed features also plays a critical role in detecting face video forgery (\textbf{Finding 5}).}

\textbf{Generation Manner and Framework.} 
This study focuses on high-quality full-image video generation techniques and predominantly adopts the diffusion-based framework. Methods~\cite{TGAN,DVD-GAN,videogpt} with poor visual video generation quality are not included in this investigation. Analysis of Figure~\ref{fig66} (b) reveals that full-image based generation techniques with diffusion framework demonstrate similar performance. Consequently, we can infer that \emph{existing diffusion-based generation techniques possess a comparable ability to generate forged videos (\textbf{Finding 6})}. Additionally, diffusion-based techniques exhibit lower performance compared to alternative methods, with face swapping yielding the best results. The potential explanation for this finding is that diffusion-based techniques, face reenactment, and face swapping alter the content of the full image, facial movements, and facial contour, respectively.   Consequently, it can be inferred that \emph{altering fewer aspects of the content leads to the generation of more authentic videos (\textbf{Finding 7})}.

\begin{figure}[!t]
  \centering
  \includegraphics[width=1\linewidth]{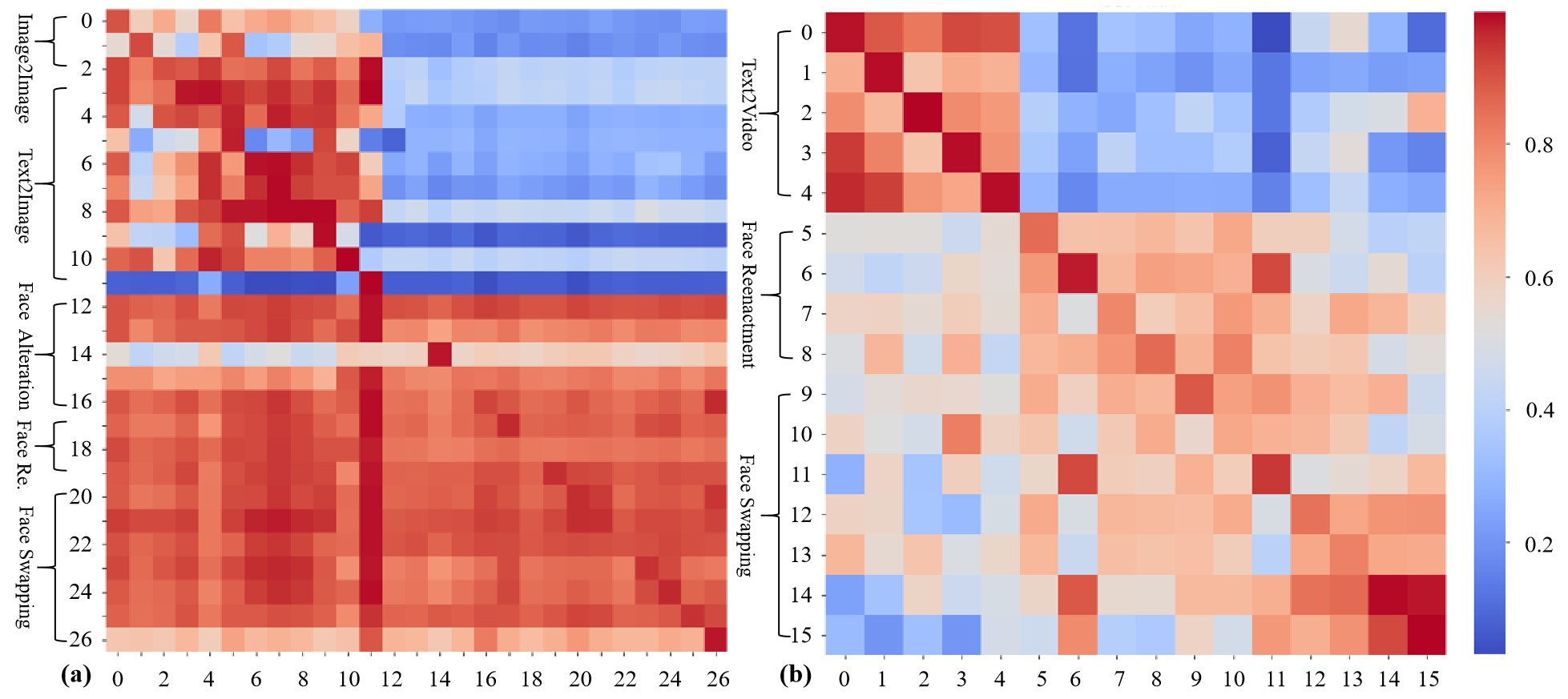}
  \caption{ 
    The cross-generalization ability verification matrices for image-level (a) and video-level (b) datasets. The training and testing samples, generated by various forgery techniques, are represented on the vertical and horizontal axes. 
    The denotation for each number is provided in the \emph{Appendix A.5}. 
  }

  \label{fig7}
\end{figure}

\subsection{Generalization Ability Evaluation to Different Forgery Techniques}
\label{Generalization}
In this section, we verify the cross-generalization ability among sub-datasets created using various forgery techniques. The results for image-level and video-level datasets, obtained using the Xception model for forgery detection, are presented in Figure ~\ref{fig7}. Furthermore, additional cross-generalization verification experiments with another $11$ forgery detection models can be found in \emph{Appendix A.5}.
\label{cross_dataset}

\textbf{Generalization Ability Across Different Forgery Techniques.} Figure ~\ref{fig7} demonstrates that models trained on localized forgery images/videos exhibit superior generalization capability than models trained on full-image forgery images/videos. This difference can be attributed to several factors. Localized forgery techniques concentrate on specific facial regions, such as eyes, mouth, and skin texture, which also serve as vital clues for detecting full-image forgery images/videos. Conversely, full-image forgery methods consider the entire image, incorporating elements like background, lighting, and environment, which introduce significant variability across different datasets. Consequently, the model's ability to generalize on localized editing samples is diminished. Thus, we can conclude that \emph{Face forgery detection methods trained on localized editing samples generally demonstrate higher generalization capability compared to those trained on full-image generation samples (\textbf{Finding 8})}.
Furthermore, from Figure ~\ref{fig7}(a), it is evident that the forgery detection technique trained and tested on samples generated by the full-image DF-GAN~\cite{DFGAN} exhibits poor and good generalization ability, respectively. This finding further confirms \emph{\textbf{Finding 3}} that full-image generation using GAN-based techniques results in low image quality, making it easily detectable by forgery detection techniques.

\textbf{Internal Generalization Ability Analysis.} 
Figure ~\ref{fig7} indicates that models trained on full-image forgery samples (Image2Image, Text2Image, and Text2Video) possess a high degree of internal generalization ability. This can be attributed to the significant similarities shared among samples generated by full-image generation techniques. Similarly, models trained on localized forgery images and videos demonstrate high and moderate internal generalization ability, respectively. Moreover, models trained on face reenactment videos and face swapping forgery videos exhibit a moderate level of generalization ability to each other. The findings imply a trend in forgery detection methods, where generalized forgery features are learned from images, while more specific forgery features are acquired from videos. This disparity may be attributed to the presence of redundant features in videos compared to single images. Hence, we can conclude that \emph{models trained on full-image forgery images, localized forgery images, and full-image forgery videos display high internal generalization ability, whereas localized forgery videos do not (\textbf{Finding 9}).}

\label{Visualization}
\begin{figure}[!t]
  \centering
  \includegraphics[width=1\linewidth]{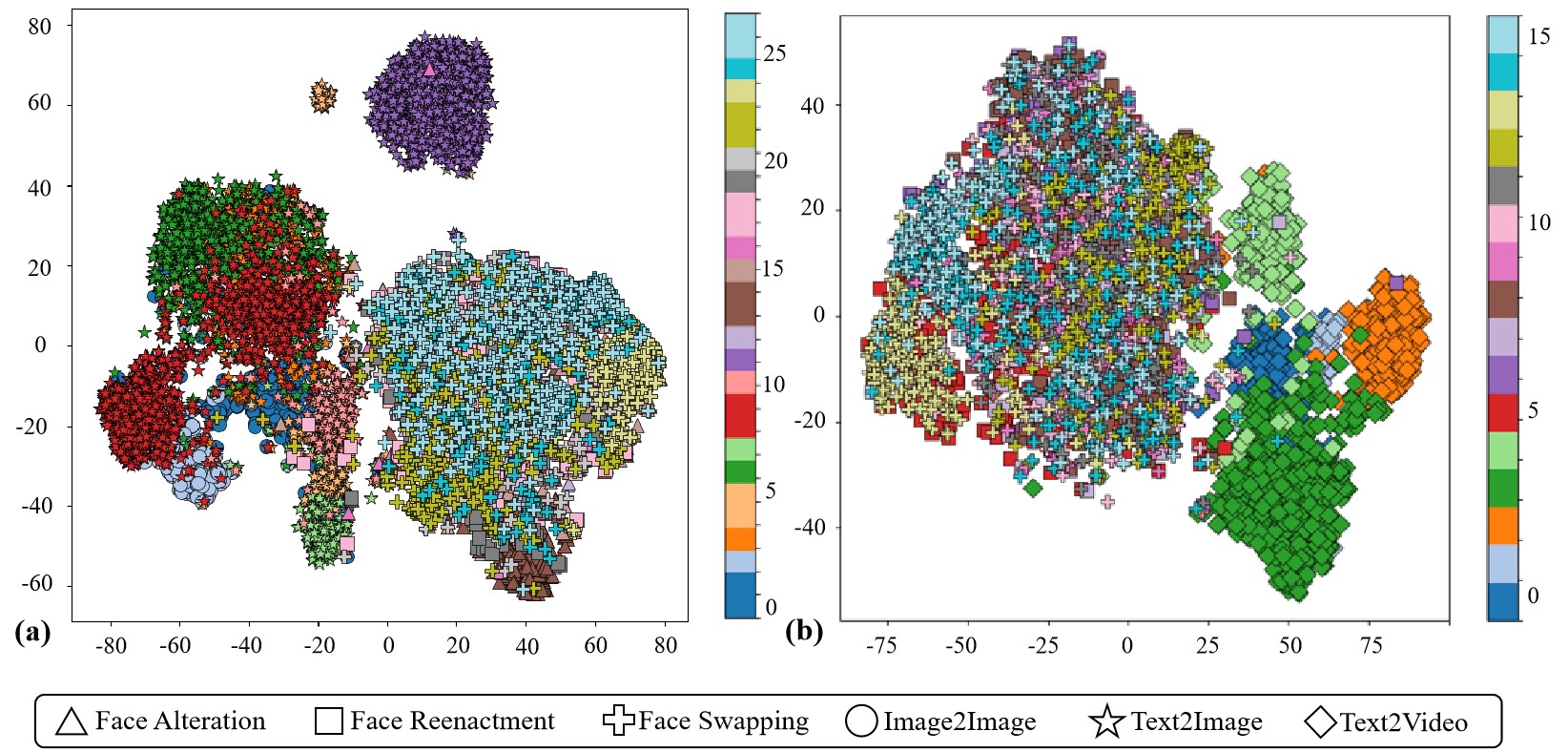} 
  \caption{ The forgery feature visualization for different forgey techniques on image-level (a) and video-level (b) datasets  with t-SNE~\cite{t-SNE}.
  }
  \label{fig8}
\end{figure}

\subsection{Visualization Analysis of Forgery Detection Features}

In this section, we utilize the fully connected layer features of the forgery detection model ResNet50~\cite{resnet} to visually evaluate the similarities among different forgery techniques. 
As illustrated in Figure~\ref{fig8}, a clear distinction is observed in the feature space between full-image forgery samples (Image2Image, Text2Image, and Text2Video) and localized editing forgery samples (face alteration, face reenactment, and face swapping). This indicates that \emph{the forgery features of full-image forgery samples and localized editing forgery samples are distinct (\textbf{Finding 10}).} It further confirms the \emph{\textbf{Finding 8\&9}}.  Further analysis is given in \emph{Appendix A.6}.

\section{Conclusion}
In this study, we present DeepFaceGen, the first comprehensive deep face forgery dataset that encompasses both localized editing and full-image generation samples. This dataset addresses the existing gap in large-scale general face forgery datasets. DeepFaceGen contains an extensive collection of over $350,000$ images and $400,000$ videos. We provide a detailed description of the dataset construction process and evaluate the performance of $13$ mainstream forgery detection techniques on samples forged using $34$ different generation techniques. By analyzing the results of these extensive experiments, we draw important findings that present novel perspectives and directions for the development of face generation and forgery detection techniques. We anticipate that this benchmark will have a far-reaching positive impact on the emerging field of artificial intelligence.

\textbf{Challenge and Future Work.} 
\label{challenge}
Based on extensive experimentation and analysis, it is evident that current forgery detection techniques suffer from drawbacks, such as low identification accuracy, poor generalization ability, and a restricted range of forgery detection types. Moreover, the rapid development of face generation techniques has created a significant discrepancy, resulting in a lag in face forgery detection. In order to address this issue, the development of a self-evolving forgery detection framework is crucial to ensure that forgery detection techniques can keep up with the advancements in face generation techniques. Additionally, this paper presents a comprehensive evaluation benchmark comprising diverse content samples, various races, and fine-grained labeling. The design of objective and comprehensive quantification metrics, as well as the establishment of a complete pipeline, are crucial for future research. Further analysis regarding challenges and future directions can be found in the \emph{Appendix A.7}. 



\appendix

\section{Appendix}

In the appendix, we provide survey of face forgery technology and face forgery detection technologies (~\ref{surveryPart1}), comprehensive statistical analysis of the DeepFaceGen dataset (~\ref{surverypart2}), and the detailed descriptions of prompts construction (~\ref{surverypart3}). We also outline the evaluation setting details (~\ref{surverypart4}), details for generalization ability verification experiments of different (~\ref{surverypart5}), and fine-grained analysis of forgery detection feature (~\ref{surverypart6}). Additionally, we give detailed challenge discussions and future directions (~\ref{surverypart7}), fine-grained attribute statistic analysis for different forgery techniques (~\ref{surverypart8}), and potential negative social impacts (~\ref{surverypart9}).

\subsection{Survey of Face Forgery Technology and Face Forgery Detection Technology}
\label{surveryPart1}

In this section, we present a comprehensive overview of both face forgery technologies and face forgery detection technologies. Regarding the former, we categorize face forgery methods into localized editing and full-image generation techniques based on their image/video generation approach. Subsequently, we discuss the forgery detection techniques designed specifically for these two types of forgery methods.

\subsubsection{Localized Editing Based Face Forgery Technology}
Localized editing based face forgery involves modifying specific facial features, such as expressions and movements. Traditional facial Photoshop (PS) techniques, which involve manual image manipulation, also fall within this scope. However, traditional PS techniques often leave detectable traces that can be identified by the naked eye. Therefore, survey of localized editing based face forgery focus on advanced deepfake methods including face swapping, face reenactment, and face alteration.

\textbf{Face Swapping.} Face swapping involves transferring the facial identity from a source image to a target image while preserving the expressions, movements, and background of the target image. Early face swapping techniques primarily relied on autoencoders. One such tool, Deepfake~\cite{r22}, popularized by Reddit users, trains the facial images of the source and target persons separately, allowing the decoder to accurately reproduce their faces. In face swapping, the encoder extracts the source person's facial features and inserts them into the target person's image using the decoder. Shaoanlu~\cite{shaoanlu2017} introduces FaceswapGAN, which employs a face swapping attention mechanism to enhance image realism. This method also addresses occlusion issues using segmentation masks. RSGAN~\cite{Natsume2018RSGANFS} is designed for face swapping using two autoencoders to represent the hair and face regions. It replaces the face's latent representation and reconstructs the image, effectively addressing issues such as mismatched face orientation and lighting. Nirkin \emph{et al}.~\cite{r27} introduces FSGAN, which uses RNN-based methods to transfer expressions and movements from the target face to the source face. FSGAN demonstrates good generalization and requires fewer training samples. Li \emph{et al}.~\cite{Li2019FaceShifterTH} introduces Faceshifter, a two-stage face-swapping method. It uses adaptive attention denormalization (AAD) for feature integration and employs a heuristic error acknowledgment refinement network (HEAR-Net) to address occlusion issues. Chen \emph{et al}.~\cite{Chen2020SimSwapAE} introduces an identity injection module to eliminate identity constraints, and enhances the loss function with weak feature matching loss to improve face synthesis quality.

\textbf{Face Reenactment.} Face reenactment preserves the target image's facial identity while replicating expressions, facial orientation, and body movements from the source image. Wang \emph{et al}.~\cite{Imaginator} introduces Imaginator, which uses a spatiotemporal feature fusion mechanism to decode continuous video from spatial features and motion. They employ two discriminators: one to evaluate the realism of facial appearances and the other to assess the realism of motions. Siarohin \emph{et al}.~\cite{Monkey-Net} introduces Monkey-Net, which separates appearance and motion information in images, enabling motion-driven animation. Monkey-Net includes a motion transfer network, an unsupervised keypoint detector, and a motion prediction network. It predicts the visual flow map for each keypoint by distinguishing keypoints in target and source images, thereby generating forged images. Siarohin \emph{et al}.~\cite{FOMM} improves on Monkey-Net by introducing local affine transformations around keypoints, which better reproduce large pose variations. Pumarola \emph{et al}.~\cite{GANimation} uses action unit annotations combined with unsupervised training and attention mechanisms to enhance model robustness. Tripathy \emph{et al}.~\cite{FACEGAN} uses action units to represent facial expressions, processing the face and background separately to improve image quality and reduce identity information leakage. CycleGAN~\cite{CycleGAN} is widely used in face reenactment due to its flexible training capabilities between source and target domains. Xu \emph{et al}.~\cite{Xu2017FaceTW} proposes a full-image reenactment method based on CycleGAN, which uses various receptive field specifications and PatchGAN to enhance image quality. Bansal \emph{et al}.~\cite{Recycle-GAN} uses CycleGAN for data-driven, unsupervised video retargeting, effectively transferring continuous information for expression-driven animation. Wu \emph{et al}.~\cite{ReenactGAN} introduces ReenactGAN, which extracts facial contours using an encoder and maps them via CycleGAN. A pix2pix generator then reconstructs the image. This method uses only feedforward neural networks, enabling real-time expression reenactment.

\textbf{Face Alteration.} Face alteration modifies specific attributes like hair color, gender, and glasses without altering facial identity. Most face alteration techniques use GAN structures. The StyleGAN series~\cite{StyleGAN1,StyleGAN2,StyleGAN3} are notable for editing facial features, while StarGAN~\cite{StarGAN1} and StarGANV2~\cite{StarGAN2} enable transformations across multiple image domains, offering better scalability. Another notable method is GANnotation~\cite{GANnotation}, which contains a triple continuity loss function for GAN-based face alteration and a direct facial expression alteration synthesis method. Kim \emph{et al}.~\cite{CAM} introduces a CAM consistency loss function based on CycleGAN's cycle consistency loss function, which helps retain feature-independent positional information and can be applied to models like StarGAN. To address scalability and diversity issues in face alteration, Li \emph{et al}.~\cite{HiSD} introduces hierarchical style disentanglement(HiSD), a hierarchical model that represents facial features as labels and attributes. Using an unsupervised approach, HiSD decouples these features, allowing for more precise modifications of target attributes.
\subsubsection{Full-image Generation Based Face Forgery Technology}
Based on the differences in network architecture, end-to-end full-image generation face forgery techniques can be categorized into gan-based models, autoregressive-based models, and diffusion-based models.
\begin{figure}[H]
  \centering
  \includegraphics[width=\linewidth]{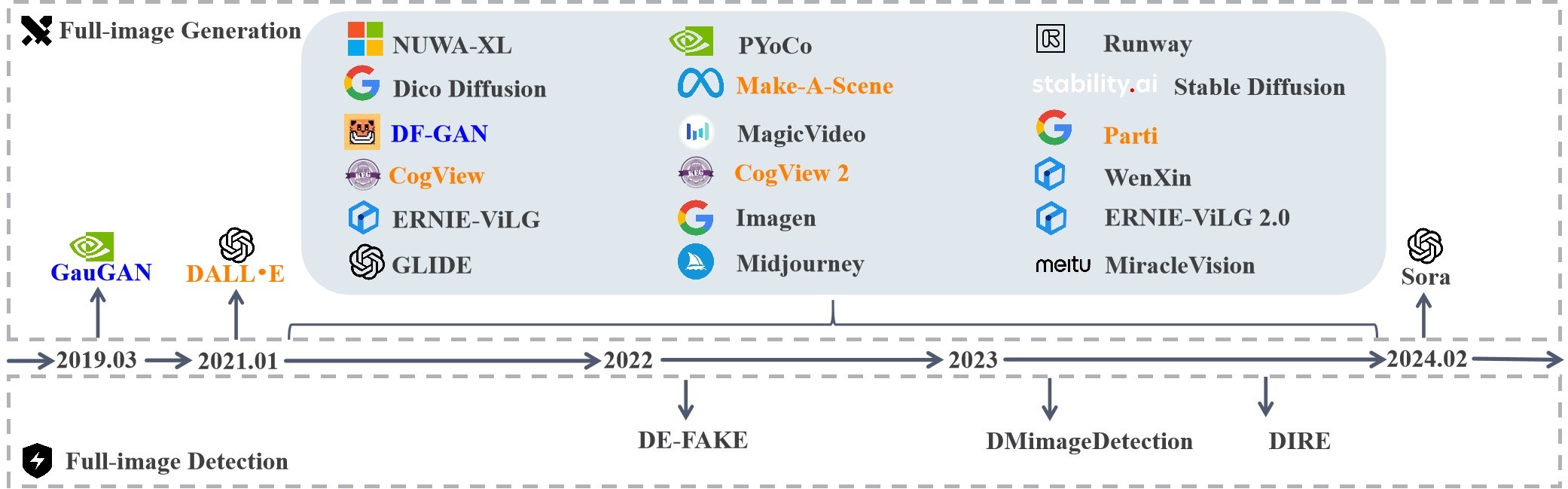} 
\caption{Full-image generation methods/products (above the timeline) and forgery detection techniques (below the timeline) are shown on a chronological timeline. 
    \textcolor{blue}{GAN}, \textcolor{orange}{Autoregressive}, and \textbf{Diffusion} are marked with \textcolor{blue}{blue}, \textcolor{orange}{orange}, and \textbf{black} fonts, respectively.}
  \label{fig1}
\end{figure}

\textbf{GAN-based Models.} Based on their model structure, GANs can be classified into single-stage generation networks and stacked architectures. DF-GAN~\cite{DFGAN}, a single-stage generation network, uses one generator, one discriminator, and a pre-trained text encoder. It maps text to images by incorporating affine transformations, enabling direct image synthesis from textual descriptions. GoGAN~\cite{GOGAN}, a stacked architecture, generates higher resolution images in stages. Each branch's generator captures the image distribution, while the discriminator assesses authenticity, refining image resolution and achieving stable training results. Despite their capabilities, GANs face stability issues and mode collapse. These limitations have led to their gradual replacement by autoregressive and diffusion models, which offer improved stability and better handling of diverse data distributions.

\textbf{Autoregressive-based Models.} Autoregressive-based models generate images by modeling spatial relationships between pixels and high-level attributes using an Encoder-Decoder architecture with a multi-head self-attention mechanism. In Text2Image generation, these models convert text and images into token sequences. The autoregressive model predicts image sequences from these tokens, which are then decoded into final images using techniques such as Variational Autoencoders (VAEs) to enhance image quality. Autoregressive models offer explicit density modeling and stable training compared to GANs. Notable examples include DALL·E~\cite{dalle2020}, which generates creative images from text prompts, CogView~\cite{ding2021cogview}, known for its high-quality image synthesis, and Make-A-Scene~\cite{Make-A-Scene}, which enables interactive image generation. However, autoregressive models face limitations in computational resources, data requirements, and training time due to their large number of parameters. Diffusion models, which offer improved efficiency and require less data, have led to a decline in interest in autoregressive models.

\textbf{Diffusion-based Models.} Diffusion-based models have become the state-of-the-art in deep generative models, surpassing previous image and video synthesis techniques. Diffusion models generate images and videos by combining noise prediction models with conditional diffusion or classifier guidance. This process allows the diffusion model to create the desired output based on the provided guidance. These models excel at handling various input conditions and mitigating mode collapse, making them dominant in fields such as Text2Image, Image2Image, Text2Video, and Image2Video synthesis. Notable examples include GLIDE~\cite{GLIDE}, known for its high-quality Text2Image generation; Imagen~\cite{Imagen}, which excels in photorealistic image synthesis; Sora~\cite{sora2024}, a state-of-the-art Text2Video model; and Stable Diffusion~\cite{stable}, which is widely used for its versatility and stability.

\subsubsection{Detection Technique for Localized Editing Based Face Forgery}
Detection techniques target localized editing based face forgeries by identifying artifacts left in various feature spaces during the forgery process. These techniques can be categorized into spatial domain-based, frequency domain-based, and temporal domain-based detection technique.

\textbf{Spatial Domain-based Detection Technique.}  Zhao \emph{et al}.~\cite{54} suggests that the key to distinguishing real from forged faces lies in subtle local details. They propose a texture enhancement module, an attention generation module, and a bi-linear attention pooling module to help the model focus on facial texture details. However, these methods often overfit to specific forgery artifacts, leading to a rapid decline in detection performance when faced with unseen forgery methods. To avoid overfitting, researchers have generated forged faces by applying certain operations to real faces. Li \emph{et al}.~\cite{55} introduces the FaceX-Ray model, which detects forgery by identifying face fusion boundaries. During training, the model predicts image authenticity and performs pixel-wise classification on the gray scale map of fusion boundaries. This method does not rely on specific forgery artifacts, showing remarkable generalization capabilities in detecting forgeries from unseen methods. Shiohara \emph{et al}.~\cite{56} argues that forgeries often contain general forgery traces. They propose Self-Blended Images (SBI), synthetic forgeries created by transforming key points within the same face image, which show strong generalization against unknown forgery methods. However, this method performs poorly against full-face synthesis methods due to its reliance on the self-forgery process.  Cao \emph{et al}.~\cite{57} introduces RECCE, combining reconstruction learning and classification to help the model learn compact features of real faces and uncover essential differences between real and fake faces. Some studies have explored the interpretability of deep face forgery detection models. Dong \emph{et al}.~\cite{58} hypothesizes that detection models identify authenticity by discerning information unrelated to facial identity. They use facial identity as an auxiliary label and designed source feature encoders and target encoders for identity recognition tasks.

\textbf{Frequency Domain-based Detection Technique.} Videos and images disseminated across online streaming media often undergo multiple compressions, resulting in low-quality images that obscure forgery artifacts. To address this issue, researchers have explored detection clues in the frequency domain. For instance,  Qian \emph{et al}.~\cite{59} finds that forgery artifacts can be effectively extracted in the frequency domain. They design a frequency-aware decomposition module to adaptively capture forgery clues within images. Additionally, they introduce a local frequency information statistics module to gather frequency information from each local region of an image and recombine these statistics into multi-channel feature maps for the frequency domain. Since artifacts appear in different regions of various images, Wang \emph{et al}.~\cite{60} introduces a multi-modal and multi-scale autoregressive model (M2TR) to detect local artifact details at different spatial levels. This model incorporates frequency domain features as auxiliary information, enhancing its capability to detect forgeries in highly compressed images. While frequency domain-based methods show strong forgery detection capabilities in highly compressed images, their performance significantly declines when encountering unknown forgery methods.

\textbf{Temporal Domain-based Detection Technique.} Temporal domain forgery detection focuses on identifying dynamic inconsistencies between video frames over time. Masi \emph{et al}.~\cite{61} proposes a dual-stream branch network. One branch extracts dynamic temporal inconsistencies from consecutive video frames, and the other amplifies artifact details using a Laplacian of Gaussian (LoG) operator. Recognizing the correlation between forgery and anomaly detection tasks, Ruff \emph{et al}.~\cite{62} introduces the deep support vector data description (Deep SVDD) loss function to improve the intra-class compactness of real faces and the inter-class distinction between real and forged faces, enhancing the model's generalization capability. Zheng \emph{et al}.~\cite{63} finds that setting the temporal convolution kernel size to 1 in 3D convolutional kernels enhances the network's ability to capture temporal inconsistencies in forged videos. However, temporal inconsistencies can be compromised by noise, compression, and other factors, leading to reduced robustness in these methods.

\subsubsection{Detection Technique for Full-image Generation Based Face Forgery}
Research achievements in the detection of full-image generation based face forgery are currently limited. Researchers are attempting to break through the mindset of searching for clues specific to localized editing based face forgery and instead seek the unique fingerprints produced by the full-image generation based face forgery process.

Sha \emph{et al}.~\cite{r7} systematically studies the detection and attribution of fake images generated by diffusion models. They compare the results of image-only input and mixed input (images and corresponding text descriptions) to explore the detection and tracing capabilities of CNN classification models.  Corvi \emph{et al}.~\cite{64} analyzes the frequency domain and model identification capabilities, concluding that diffusion-generated images have unique fingerprints similar to GAN images. Wang \emph{et al}.~\cite{DIRE} find that the diffusion reconstruction effect of fake images is superior to that of real images. They use the difference between the reconstructed image and the original image, called Diffusion Reconstruction Error (DIRE), for binary classification to determine authenticity, showing higher generalization ability. However, these methods are tested on small, self-created datasets, and their experimental conclusions lack generality. Additionally, they do not specifically focus on detecting face forgeries. Currently, the detection of faces generated by diffusion models remains relatively unexplored.

\subsection{DeepFaceGen Detailed Statistical Data}
\label{surverypart2}
In order to construct a robust and extensive benchmark for the detection of face forgery, we carefully consider a range of critical factors including the manner of generation, generation framework, content diversity, ethnic fairness, and label richness throughout the benchmark development process. Following this, we provide detailed introduction to the forged face samples and authentic face samples in DeepFaceGen.

\begin{figure}[!t]
  \centering
  \includegraphics[width=\linewidth]{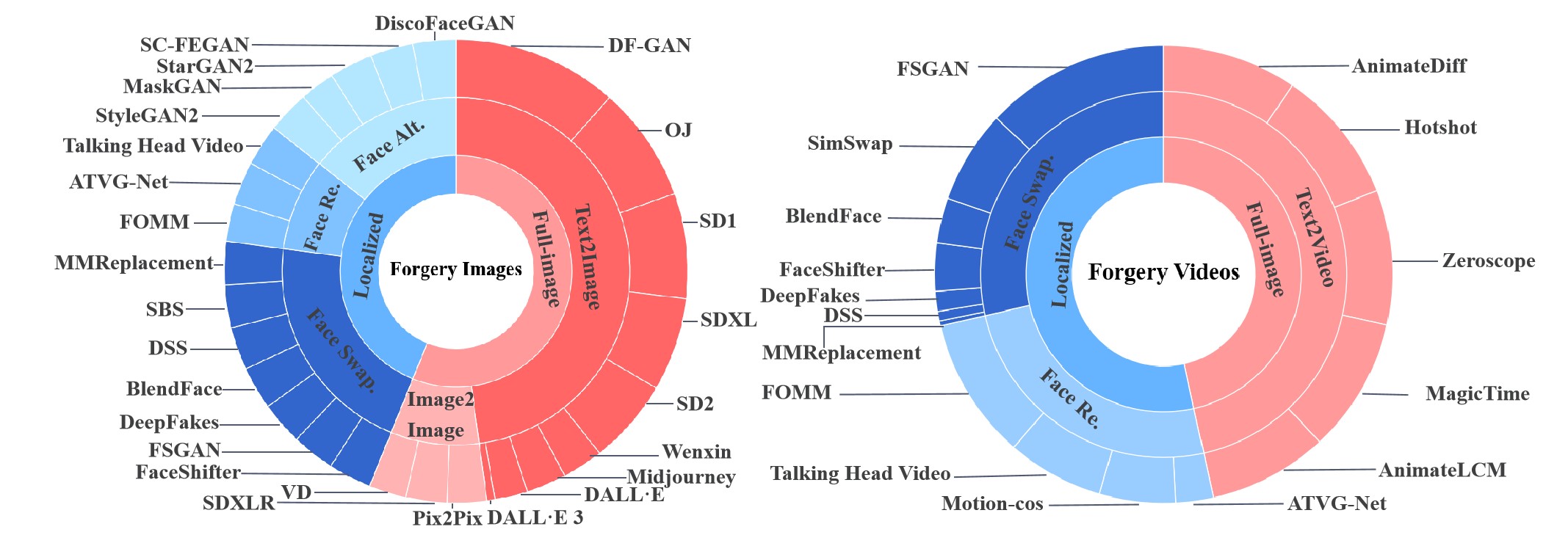} 
  \caption{
Composition and porportion illustration of image- and video-level sets. At the image-level, DeepFaceGen utilizes $27$ face forgery methods. At the video-level, it employs $16$ methods. In both levels, the forged data maintains an approximate balance between localized editing based face forgery technology and full-image generation based face forgery technology.}
  \label{fig4}
\end{figure}

\textbf{Forged Face Samples}. The forged face samples of DeepFaceGen consists of $34$ types of forgery methods. The number of forged images/videos reaches $350,264$/$423,548$. For content diversity, we collected $143,579$ forged images and $93,497$ forged videos from~\cite{r13} and~\cite{r19}. As shown in Figure~\ref{fig4}, the forged images contain $27$ forgery methods, including localized editing based and full-image based generation. Forged samples between both generation methods are roughly balanced. The localized editing based samples include face swapping, face reenactment and face alteration. In the full-image based generation, sufficient Text2Image and Image2Image samples are generated according to the input modality. At the video-level, a rough balance is similarly maintained between the samples generated by the $16$ forgery methods. In the process of generating forged video/image samples, in order to maintain ethnic fairness, we control the balance of skin color through text prompt in full-image based generation. Localized editing based samples also fit ethnic fairness by employing SkinToneClassifier~\cite{skin}. Additionally, we employ YOLO~\cite{yolo} with manual screening to eliminate low-quality data. The detailed forged statistical data can be seen in Table~\ref{DeepFaceGen_Detailed_Statistical_Data}.

\textbf{Authentic Face samples.} In order to ensure content diversity and ethnic fairness in the authentic face samples used in DeepFaceGen, we obtained real samples from reputable sources including ~\cite{r13}, ~\cite{r19}, ~\cite{CN-CVS}, and ~\cite{CMLR}. Specifically, we collected $482$ and $463,101$ real images from~\cite{r13} and~\cite{r19}, and $19,942$, $590$, $99,630$, $193,245$ real videos from ~\cite{CMLR},~\cite{r13}, ~\cite{r19}, and~\cite{CN-CVS}.
The final collection consists of $463,583$ images and $313,407$ videos, encompassing diverse ages, genders, skin tones, expressions, hair styles, hair colors, backgrounds, dressing styles, and glasses.

\begin{table*}[!t]
    \centering
    \caption{Detailed Statistical Data of DeepFaceGen.}
    \label{DeepFaceGen_Detailed_Statistical_Data}
    \begin{tabular}{c|c|c|c|c|c}
        \toprule
        \textbf{Manner}&\textbf{Subset} & \textbf{Methods} & \textbf{Images} & \textbf{Videos} & \textbf{Labels} \\ 
        \midrule
        \multirow{17}{*}{\rotatebox{90}{Localized editing}}& \multirow{8}{*}{Face Swapping} 
         & FaceShifter & \numprint{10500} & \numprint{14387} & \multirow{8}{*}{n-way labels} \\ 
         && FSGAN & \numprint{10500} & \numprint{55205} &  \\ 
         && DeepFakes & \numprint{10500} & \numprint{6000} &  \\ 
         && BlendFace & \numprint{10500} & \numprint{13491} &  \\ 
         && DSS & \numprint{10500} & \numprint{2866} &  \\ 
         && SBS & \numprint{10500} & - &  \\ 
         && MMReplacement & \numprint{10500} & \numprint{1461} &  \\ 
         && SimSwap & - & \numprint{27786} &  \\ \cmidrule(r){2-6}
        &\multirow{4}{*}{Face Reenactment} 
         & Talking Head Video & \numprint{9203} & \numprint{28935} & \multirow{4}{*}{n-way labels} \\ 
         && ATVG-Net & \numprint{10500} & \numprint{11273} &  \\ 
         && Motion-cos & - & \numprint{22811} &  \\ 
         && FOMM & \numprint{10235} & \numprint{42411} &  \\ \cmidrule(r){2-6}
        &\multirow{5}{*}{Face Alteration} 
         & StyleGAN2 & \numprint{10263} & - & \multirow{5}{*}{n-way labels} \\ 
         && MaskGAN & \numprint{8613} & - &  \\ 
         && StarGAN2 & \numprint{10500} & - &  \\ 
         && SC-FEGAN & \numprint{10500} & - &  \\ 
         && DiscoFaceGAN & \numprint{10500} & - &  \\ \midrule
        \multirow{17}{*}{\rotatebox{90}{Full-image}}&\multirow{9}{*}{Text2Image}
         & OJ & \numprint{28203} & - & \multirow{9}{*}{\shortstack[l]{n-way labels \\ prompt labels}} \\ 
         && SD1 & \numprint{25677} & - &  \\ 
         && SD2 & \numprint{20898} & - &  \\ 
         && SDXL & \numprint{22839} & - &  \\ 
        & & Wenxin & \numprint{9989} & - &  \\ 
         && Midjourney & \numprint{9784} & - &  \\ 
         && DF-GAN & \numprint{40320} & - &  \\ 
         && DALL·E & \numprint{8000} & - &  \\ 
         && DALL·E 3 & \numprint{2000} & - &  \\ \cmidrule(r){2-6}
        &\multirow{6}{*}{Text2Video}
         & AnimateDiff & - & \numprint{40320} &  \multirow{5}{*}{\shortstack[l]{n-way labels \\ prompt labels}}  \\ 
         && AnimateLCM & - & \numprint{35642} &  \\ 
         && Hotshot & - & \numprint{40320} &  \\ 
         && Zeroscope & - & \numprint{40320} &  \\ 
         && MagicTime & - & \numprint{40320} &  \\ \cmidrule(r){2-6}
        &\multirow{3}{*}{\shortstack[l]{Image2Image}} 
         & Pix2Pix & \numprint{9620} & - &  \multirow{3}{*}{\shortstack[l]{n-way labels \\ prompt labels}}  \\ 
         && SDXLR & \numprint{9990} & - &  \\ 
        & & VD & \numprint{9130} & - &  \\ \midrule
        \multicolumn{3}{c|}{\textbf{Total}} & \textbf{\numprint{350264}} & \textbf{\numprint{423548}} &  \\ 
        \bottomrule
    \end{tabular}
\end{table*}
\subsection{Detailed Descriptions of Prompts Construction}
\label{surverypart3}
\begin{figure}[!t]
  \centering
\includegraphics[width=1.0\linewidth]{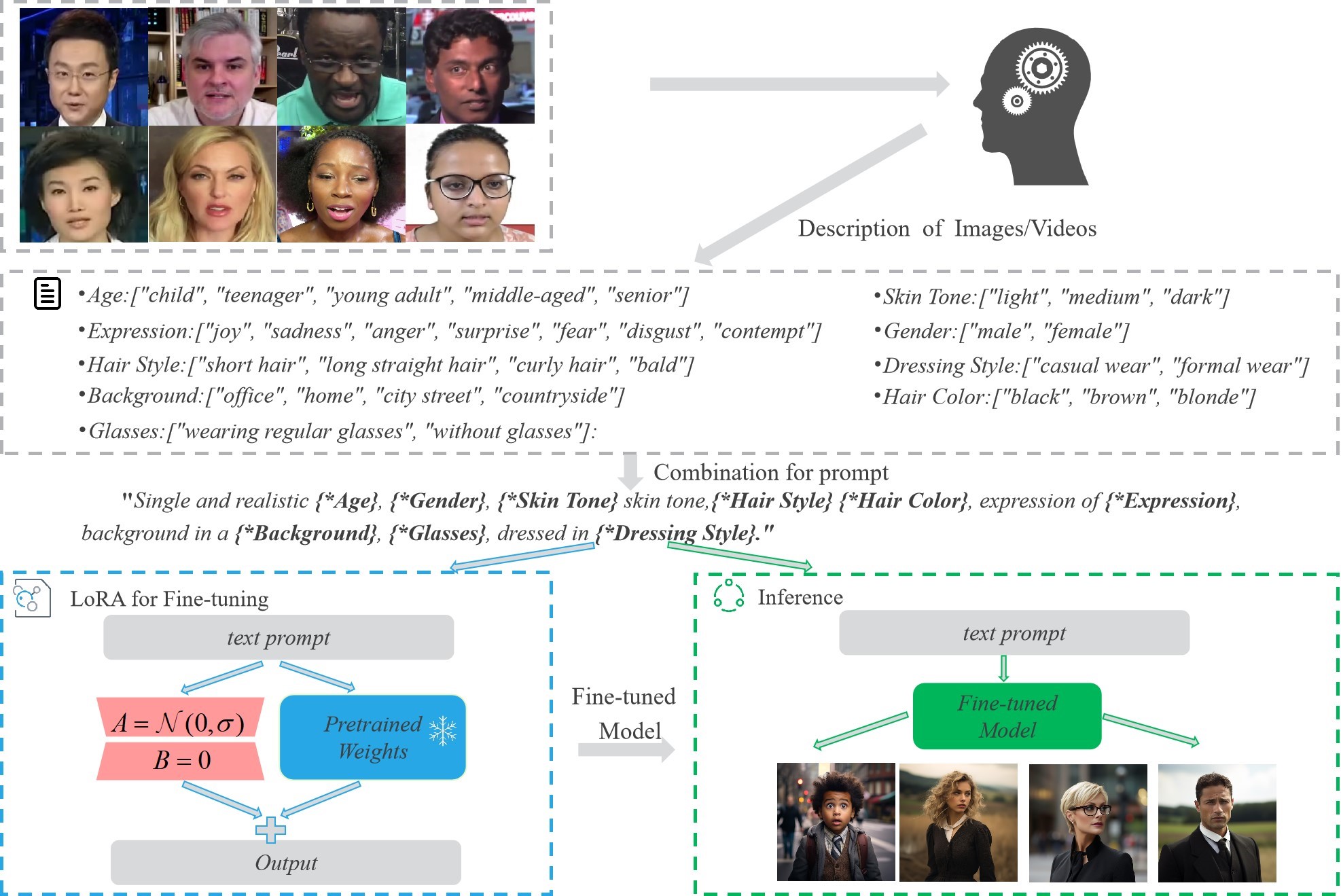} 
  \caption{Pipeline of prompts construction. It consists of four parts: the establishment of face description information, the construction of description attributes, the fine-tuning of pre-trained models and the generation of forged samples. After establishing comprehensive attributes to describe face information from images and videos, rich and comprehensive text prompts can be obtained by iterating the combination of description attributes. Then, LoRA\cite{hu2022lora} is used to fine-tune the generative model to the field of face generation for the final generation task.}
  \label{fig2}
\end{figure}

 In the design of prompts, we strive to achieve both content diversity and fairness, which are accompanied by a strong emphasis on detailed prompt descriptions. Following this, we designed a complete expressive framework for each prompt sentence based on the face information that humans take into account when describing faces. The prompt sentence framework contains $9$ description attributes: ages, genders, skin tones, expressions, hair styles, hair colors, backgrounds, dressing styles, and glasses. Each description attribute contains a detailed scenario situation. By iterating through the combination of $9$  attributes, we can generate over $40,000$ prompts. This design ensures data balance across the various text attributes. Then, we use LoRA~\cite{hu2022lora} to fine-tune the selected pretrained model and generate forged samples fine-tuned with deepfake samples. The detailed pipeline of prompts construction is shown in Figure~\ref{fig2}.

\subsection{Evaluation Details}
In this section, we provide a detailed introduction to the selected forgery detection methods and disclose the implementation details during the experimental process.
\label{surverypart4}
\subsubsection{Forgery Detection Models}
Following the basic backone used by the $13$ forgery detection methods, we introduce the forgery detection methods in detail.
\begin{itemize}
\item\textbf{MesoNet}~\cite{Afchar2018MesoNetAC} is a face forgery detection algorithm based on mid-level information from image noise. This approach effectively addresses the challenges of diminished image noise and the difficulty of distinguishing forged video frames using high-level semantic features. Its shallow architecture enhances sensitivity to medium and large-scale features, thereby improving the capability of detecting facial characteristics.

\item\textbf{Xception}~\cite{Xception} is a convolutional neural network architecture entirely based on depthwise separable convolution layers, simplifies the decoupling of channel correlation and spatial correlation to derive depthwise separable convolutions. This enables efficient extraction of complex features from images and video frames.

\item\textbf{EfficientNet-B0}~\cite{tan2020efficientnet} is the baseline network of the
EfficientNet family, which is developed by leveraging a multi-objective neural architecture search based
on mobile inverted bottleneck MBConv~\cite{MBConv} with
squeeze-and-excitation optimization~\cite{squeeze} added to it.

\item\textbf{F3-Net}~\cite{F3NET} utilizes two complementary frequency-aware cues: frequency-aware decomposed image components and local frequency statistics. These cues are deeply explored through a dual-stream collaborative learning framework to detect subtle forgery patterns. 

\item\textbf{RECCE}~\cite{recce} is a reconstruction and classification learning framework designed to learn common characteristics of real faces by reconstructing face images. It trains a reconstruction network using real face images and employs the latent features of this network to classify real and forged faces. Due to the inconsistency in data distribution between real and forged faces, the reconstruction errors for forged faces and can accurately highlight the forged regions.

\item\textbf{DNANet}~\cite{dnaet} adopts
pre-training on image transformation classification and
patchwise contrastive learning to capture globally consistent features that are invariant to semantics. It can focus on architecture-related traces and strengthen the global consistency of extracted features.

\item\textbf{FreqNet}~\cite{tan2024frequencyaware} is a lightweight frequency space learning network designed for generalizable forgery image detection. This approach leverages the power of frequency domain learning, providing an adaptable solution for the challenging problem of deepfake detection across diverse sources and GAN models. The methodology includes practical and compact frequency learning plugin modules that integrate with CNN classifiers to enable them to operate effectively within the frequency domain.

\item\textbf{CViT}~\cite{cvit} is a model composed of two main components: Feature Learning (FL) and the Vision Transformer (ViT). The FL component, a stack of convolutional operations without a fully connected layer, extracts features from face images. These features are then processed by the ViT, which converts them into a sequence of image pixels for detection. 

\item\textbf{SLADD}~\cite{sladd} aims to generalize well in unseen scenarios. It operates on the principle that a generalizable detector should be sensitive to various types of forgeries. SLADD enriches the diversity of forgeries by synthesizing augmented forgeries using a pool of forgery configurations and enhances sensitivity by training the model to predict these configurations. 

\item\textbf{Exposing}~\cite{ba2024exposing} is an information bottleneck-based framework for deepfake detection that aims to extract broader forgery clues. It captures a wide range of forgery clues by extracting multiple non-overlapping local representations and fusing them into a global, semantically rich feature. 
\end{itemize}
\subsubsection{Implementation Details}
\textbf{Preproccess.} The image and video datasets are divided into training, validation, and test subsets in a ratio approximately $7:1:2$. To ensure fairness in evaluation, each subset maintains a ratio of real to fake instances close to $1 : 1$. For video-level
evaluations, the video files in the dataset need to be extracted and stored as individual video frames. Given the varying lengths of the video files we collected and generated, we standardize the number of frames extracted from each video to $24$. Additionally, since the authors of SLADD~\cite{sladd} did not disclose the process for creating masks, we adopted the following approach: the mask for real data is set to an all-zero matrix, indicating that there are no forgery regions in the input image. For forged data, we use YOLO~\cite{yolo} to obtain the face bounding box, and then convert the bounding box into a binary mask image, with the forgery region set to $1$ and all other areas set to $0$.

\textbf{Training.}  We all follow the original hyperparameter settings in the evaluation methods. The loss function for SLADD~\cite{sladd} is set to MSE, while the loss functions for MesoNet~\cite{Afchar2018MesoNetAC}, EfficientNet-B0~\cite{tan2020efficientnet}, Xception~\cite{Xception}, F3-Net~\cite{F3NET}, DNANet~\cite{dnaet}, RECCE~\cite{recce}, and CViT~\cite{cvit} are set to CrossEntropyLoss. In particular, based on CrossEntropyLoss, Exposing~\cite{ba2024exposing} designed the local information loss based on the theoretical analysis of mutual information to ensure the orthogonality and adequacy between local features. The optimizer for all models is Adam with a learning rate of $1 \times 10^{-5}$. The batch size is set to $128$. All models are pre-trained on ImageNet. All images in the dataset were resized to a fixed resolution of $299 \times 299$ pixels and normalized to have pixel values in the range [$0$, $1$]. 

\textbf{Inference.} We only perform single-crop inference, and directly scale the input face image to the input spatial size of the model.

\begin{sidewaystable}[!thbp]
\centering
\caption{The AUC scores of Cross-generalization Ability Verification Experiments (D: Detection technique, F: Forgery method).}
\label{table4}
 \begin{adjustbox}{width=1\textwidth}
\begin{tabular}{cccccccccccc}
\hline
\diagbox{D}{F}&Year& Modality &MMReplacement& FaceShifter & FSGAN & DeepFakes & BlendFace & SBS & DSS & ATVG-Net & FOMM \\
\hline
Xception & 2019 &Image&0.758	&0.798	&0.785	&0.775	&0.767	&0.764	&0.769	&0.805	&0.752
\\
EfficientNet-B0 & 2019 &Image&0.701	&0.751	&0.714	&0.698	&0.700	&0.692	&0.700	&0.713	&0.703
\\
F3-Net & 2020&Image &0.785	&0.894	&0.865	&0.790	&0.806	&0.785	&0.803	&0.866	&0.788
\\
RECCE & 2022 &Image& 0.799	&0.892	&0.855	&0.794&0.796	&0.788	&0.794	&0.849	&0.794
\\
DNANet & 2022 &Image& 0.769	&0.811	&0.799	&0.783	&0.783	&0.774	&0.782	&0.815	&0.764
 \\
FreqNet & 2024 &Image& 0.768	&0.882	&0.848	&0.780	&0.771	&0.773	&0.780	&0.857	&0.768
\\
\hline
\diagbox{D}{F} &Year&Modality & Talking Head Video &StarGAN2&StyleGAN2&MaskGAN&SC-FEGAN&DiscoFaceGAN&DALL·E 	&DALL·E3 	&Wenxin  \\
\hline
Xception & 2019&Image &0.738&0.783	&0.744	&0.730	&0.754	&0.720	&0.791	&0.820	&0.748	\\
EfficientNet-B0 &2019&Image&0.693& 0.679	&0.690	&0.687	&0.705	&0.663	&0.725	&0.805	&0.701\\
F3-Net & 2020 &Image&0.764 &0.794	&0.773	&0.785	&0.792	&0.727	&0.862	&0.764	&0.769\\
RECCE & 2022 &Image&0.762& 0.788	&0.774	&0.780	&0.798	&0.730	&0.875	&0.864	&0.803\\
DNANet & 2022&Image &0.750& 0.788	&0.756	&0.745	&0.766	&0.729	& 0.791	&0.857	&0.757\\
FreqNet & 2024 &Image&0.752&0.771	&0.755	&0.759	&0.771	&0.726	&0.824	&0.836&0.820\\
\hline
\diagbox{D}{F}& Year &Modality & SD1	&OJ	&SD2	&SDXL	&DF-GAN	&Midjourney&SDXLR	&pix2pix	&VD  \\
\hline
Xception & 2019&Image &0.820	&0.844	&0.842	&0.799	&0.924	&0.806	&0.751	&0.696	&0.709\\
EfficientNet-B0 & 2019&Image &0.743	&0.761	&0.698	&0.737	&0.697	&0.823	&0.683	&0.558	&0.689\\
F3-Net & 2020 &Image&0.846	&0.852	&0.856	&0.670	&0.930	&0.855	&0.824	&0.758	&0.857\\
RECCE & 2022 &Image& 0.871	&0.866	&0.870	&0.755	&0.937	&0.793	&0.815	&0.714	&0.862\\
DNANet & 2022 &Image&0.832	&0.856	&0.856	&0.809	&0.873	&0.805	&0.766	&0.695	&0.766\\
FreqNet  & 2024&Image &0.900&0.927	&0.907	&0.794	&0.976	&0.820	&0.776	&0.738	&0.831 \\
\hline
\diagbox{D}{F}&Year& Modality&Talking Head Video	&FSGAN	&DeepFakes	&BlendFace	&DSS	&MMReplacement	&SimSwap	&FaceShifter	&ATVG-Net\\
\hline
MesoNet	&2018&Video	&0.423	&0.477	&0.460	&0.411	&0.818	&0.523	&0.589	&0.489	&0.577\\
EfficientNet-B0	&2019&Video	&0.531	&0.624	&0.589	&0.713	&0.882	&0.563	&0.621	&0.677	&0.656\\
Xception	&2019&Video	&0.711	&0.594	&0.608	&0.681	&0.893	&0.642	&0.673	&0.652	&0.663\\
F3-Net	&2020&Video	&0.682	&0.742	&0.519	&0.633	&0.873	&0.682	&0.467	&0.605	&0.693\\
CViT	&2021&Video	&0.773	&0.612 	&0.593	&0.580	&0.842	&0.563	&0.593	&0.736	&0.642\\
SLADD	&2022&Video	&0.773	&0.643	&0.569	&0.761	&0.875	&0.683	&0.549	&0.712	&0.751\\
Exposing	&2024&Video	&0.683	&0.613	&0.588	&0.720	&0.916	&0.653	&0.684	&0.782	&0.653\\
\hline
\diagbox{D}{F}&Year&Modality	&Motion-cos&	FOMM	&AnimateDiff	&AnimateLCM	&Hotshot	&Zeroscope	&MagicTime&-&-\\
\hline
MesoNet	&2018&Video	&0.443	&0.582	&0.565	&0.639	&0.752	&0.801	&0.781&-&-\\
EfficientNet-B0	&2019&Video	&0.558	&0.672	&0.532	&0.704	&0.864	&0.807	&0.801&-&-\\
Xception	&2019&Video	&0.699	&0.458	&0.794	&0.763	&0.759	&0.857	&0.821&-&-\\
F3-Net	&2020&Video	&0.650	&0.591	&0.688	&0.746	&0.656	&0.761	&0.732&-&-\\
CViT	&2021&Video	&0.663	&0.716	&0.801	&0.743	&0.804	&0.798	&0.735&-&-\\
SLADD	&2022&Video	&0.643	&0.685	&0.769	&0.784	&0.828	&0.805	&0.823&-&-\\
Exposing	&2024&Video	&0.710	&0.593	&0.854	&0.837	&0.897	&0.853	&0.836&-&-\\
\hline
\end{tabular}
\end{adjustbox}
\end{sidewaystable}
\begin{figure}[!t]
  \centering
  \includegraphics[width=\linewidth]{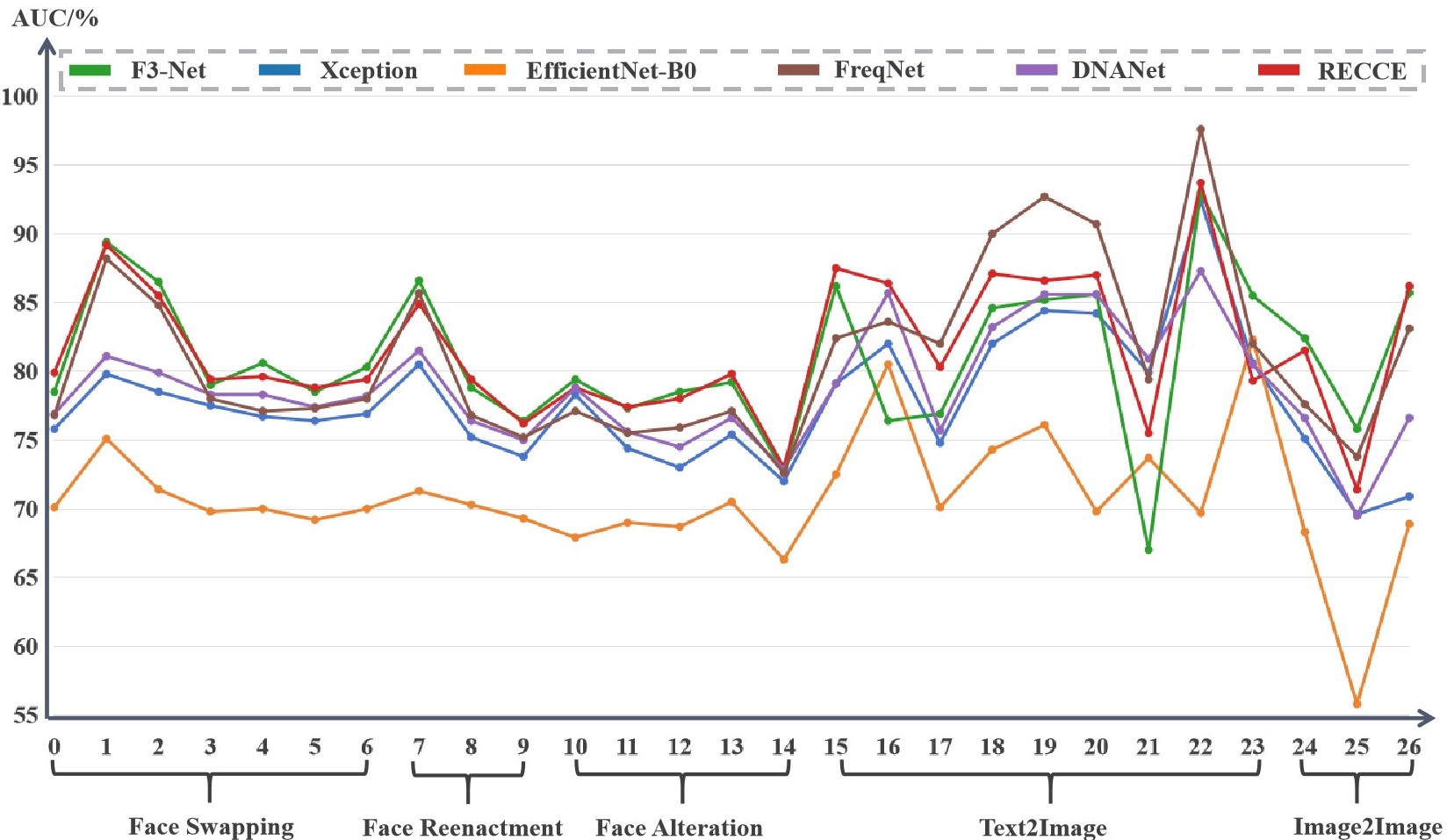} 
\caption{The cross-generalization ability comparison for various image-level forgery detection methods. The horizontal axes represent $5$ categories of image forgery techniques. All forgery detection methods are trained on the FaceShifter subset, which has demonstrated the best generalization performance among the detection techniques described in the main manuscript. These methods are subsequently tested using samples generated by the aforementioned forgery techniques.
}
  \label{zhuxing_image}
\end{figure}
\begin{figure}[!t]
  \centering
  \includegraphics[width=\linewidth]{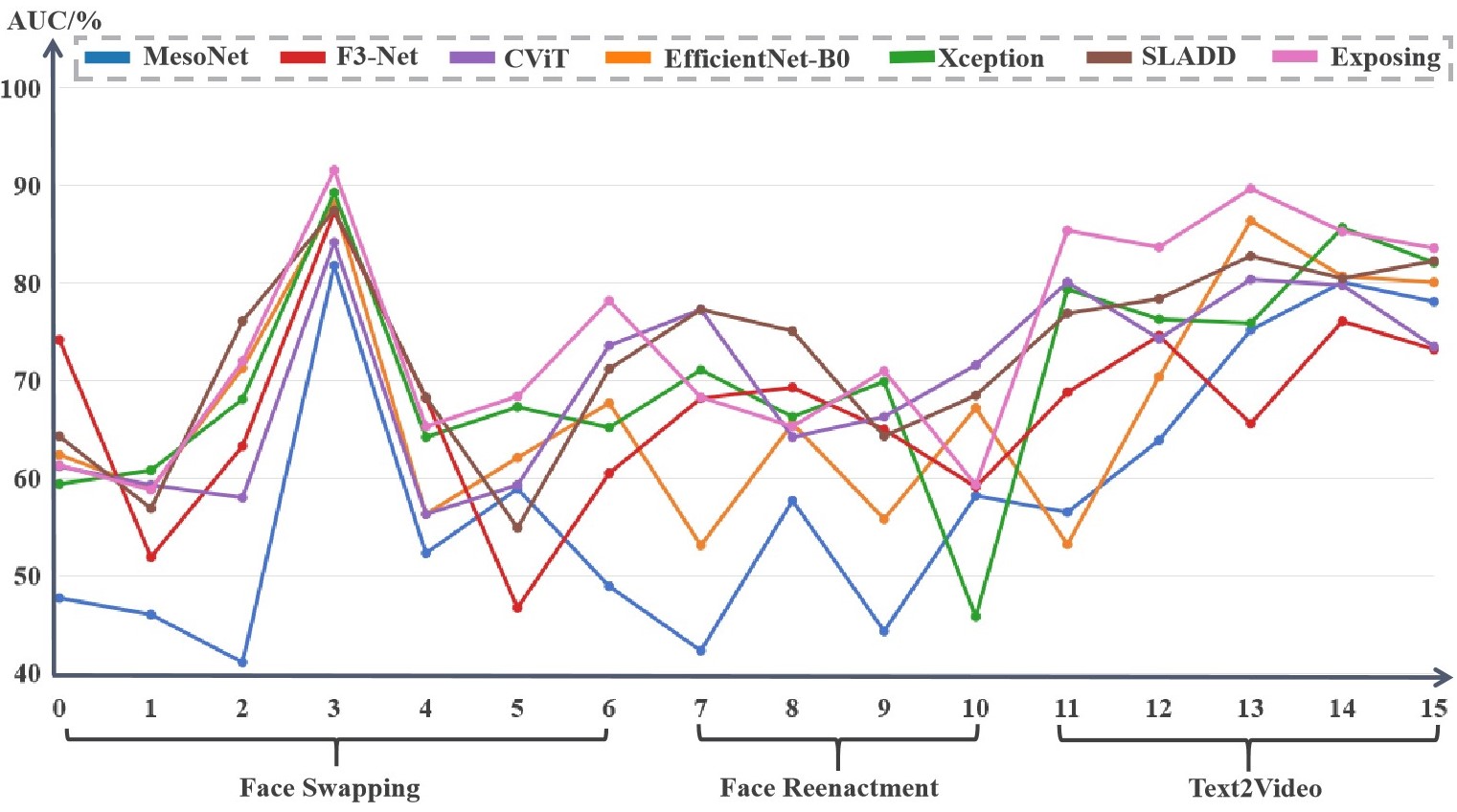} 
\caption{
The cross-generalization ability comparison for various video-level forgery detection methods. The horizontal axes represent $3$ categories of localized editing-based video forgery techniques. 
All forgery detection methods are trained on the DSS subset, chosen for its superior generalization performance among the detection techniques described in the main manuscript. Subsequently, these methods are tested using samples generated by the aforementioned video-level forgery techniques.
}
  \label{zhuxing_video}
\end{figure}
\subsection{Details for Cross-generalization Ability Verification Experiments}
\label{surverypart5}
In this section, we employ $13$ forgery detection methods to evaluate the cross-generalization capabilities among sub-datasets. The forgery detection methods are first trained on the subsets that exhibited the best generalization performance in the broad capability evaluation experiments of different forgery techniques discussed in the main text ( FaceShifter subset at the image level and DSS subset at the video level). Subsequently, the generalization performance is tested across various subsets. As shown in Figure~\ref{zhuxing_image} and Figure~\ref{zhuxing_video}, models with detail extraction modules, such as Exposing~\cite{ba2024exposing}, FreqNet~\cite{tan2024frequencyaware} and RECCE~\cite{recce}, achieve higher evaluation metrics for identifying editing forged data, which corresponds to \textbf{\emph{Finding 1}}. During the generalization test from localized editing forgery to full-image generation forgery, it is easier to detect data generated by DF-GAN, further validating \textbf{\emph{Finding 3}}. Additionally, when using localized editing forgery images/videos as training data, the internal generalization ability of video forgery detection models is significantly lower than that of image forgery detection models, further confirming \textbf{\emph{Finding 9}}. The detailed experimental results can be viewed in Table~\ref{table4}.
\begin{figure}[!t]
  \centering
  \includegraphics[width=\linewidth]{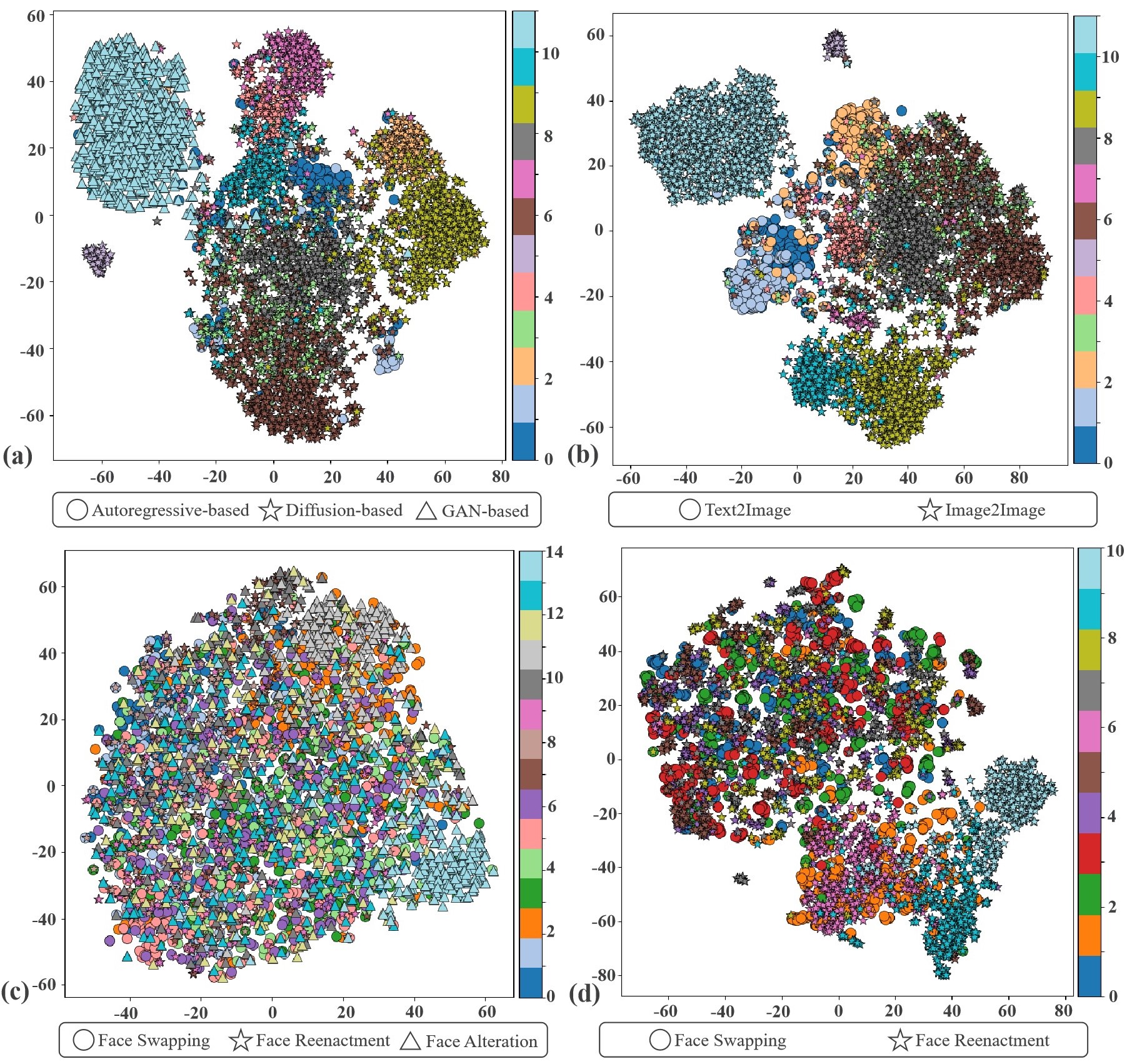} 
\caption{ The forgery feature visualization for different forgery techniques on image-level (a-c) and
video-level (d) datasets with t-SNE~\cite{t-SNE}. (a) different generation frameworks, (b) different input modalities, (c) and (d) different generation manners. }
  \label{tsne_buchong}
\end{figure}
\subsection{Fine-grained Analysis of Forgery Detection Feature}
\label{surverypart6}
As shown in Figure~\ref{tsne_buchong}, we conduct a fine-grained visual analysis of forgery detection features. Based on Figure~\ref{tsne_buchong} (a), it is evident that the forgery features of GAN-based model are significantly different from those of Diffusion-based and Autoregressive-based models. This phenomenon provides an explanation for \textbf{\emph{Finding 3}} from the perspective of feature distribution. In Figure~\ref{tsne_buchong} (b), the forgery feature distributions are similar when using text and image as input modalities, which corresponds to \textbf{\emph{Finding 4}}. Additionally, Figures~\ref{tsne_buchong} (c) and (d) demonstrate that \emph{the forgery features of localized editing techniques do not show significant differences between images and videos \textbf{(Finding 11)}}.

\subsection{Challenges and Future Work}

In light of the rapid advancements in face generation techniques, the progress of face forgery detection techniques has significantly lagged behind. Extensive experimentation and analysis reveal several deficiencies in the current forgery detection methods, including inadequate identification accuracy, limited generalization capabilities, and restricted scope for detecting various types of forgery. This section provides a comprehensive overview of the existing challenges in face forgery detection and offers potential valuable directions for future research.

\label{surverypart7}
\subsubsection{Challenges}
\begin{itemize}
  \item  \textbf{Difficulty in Handling Complex Scenarios.} The diversity of complex scenarios increases the difficulty of face forgery detection tasks. Real-world face forgery detection can be affected by environmental factors such as changes in lighting conditions, which can alter shadows and highlights on the face, making it appear darker or brighter. Changes in camera angles can distort facial shapes and features, making the face look twisted or misaligned. Additionally, variations in background complexity can blur the edges of the face or blend it with the background, making it appear unclear or disproportionate. These factors can impact the authenticity and reliability of detection results, increasing the difficulty of recognizing and detecting forgeries.
  \item \textbf{Poor Generalization Performance.} Although current detection models perform well on individual face forgery datasets, their generalization across different datasets remains inadequate. In real-world scenarios, the type of face forgery method used is often unknown, making it difficult to determine the specific type of forgery. Therefore, using pre-trained face forgery detection models for real-world tasks may result in unreliable detection outcomes.
  \item \textbf{Oversimplified Forgery Detection Tasks.} Current face forgery detection tasks focus primarily on binary classification of whether the content is forged, which is relatively crude. In real-world scenarios, there is often a need for tracing the source of the forgery, which is crucial for determining responsibility and uncovering the truth. In face video forgery tasks, attackers often target only a few video frames or audio segments to alter the video content. However, forgery detection models that focus on video-level forgery detection can easily overlook the characteristics of forged segments, significantly increasing the likelihood of detection errors.
\end{itemize}
\subsubsection{Future Work}
\begin{itemize}
  \item \textbf{Objective Quantification of Evaluation Benchmarks.} With the increasingly complex and realistic content forgery scenarios brought about by the development of AIGC technologies, current evaluation benchmarks rely on specific model performance metrics, which can be limiting. In real-world scenarios, designing evaluation benchmarks that can accurately quantify the multi-angle forgery detection capabilities and even the adaptability of models is a crucial direction for future exploration.
  \item  \textbf{Dynamic Updating of Benchmark Data.} When designing evaluation benchmarks, it is essential to consider the existence of diverse face forgery types. Regularly updating benchmark datasets to include the latest forgery techniques can help the benchmarks stay close to the complex real-world scenarios. Integrating user feedback data can provide new ideas for dynamically updating benchmark datasets. Additionally, as deep forgery technologies continue to evolve, establishing a dynamic labeling mechanism to address new deep forgery techniques and generative models is becoming increasingly important.
  \item \textbf{Building General Forgery Detection Scenarios.} Although we have constructed a general face deep forgery detection dataset that includes both localized editing based and full-image generation based face forgery techniques, incorporating both image and video modalities, the audio aspect remains a gap. Furthermore, given the relatively unexplored state of detecting face forgeries generated by diffusion methods, designing general forgery detection techniques based on the inherent differences between real and forged videos, as well as the local feature similarities and model inference paths, is a critical issue that needs to be addressed in the coming years.
  \item \textbf{Emphasis on Robustness of Forgery Detection Models.} The robustness of forgery detection models is key to maintaining stability and reliability in real-world scenarios with complex and variable content. Introducing adversarial samples during training and testing can enhance the robustness of models. However, while adding noise and adversarial samples can improve robustness to some extent, it can also lead to a loss in detection performance. Exploring the inherent characteristics of real samples to identify differences between forged and real samples and developing detection methods that can handle any face forgery product while ensuring detection accuracy is a primary research direction for the future.
  \item \textbf{Self-Evolving Forgery Detection Frameworks.} Forgery techniques and forgery detection techniques are mutually aligned and promote each other. Forgery technologies generally advance faster than forgery detection technologies, leading to significant harm from forged face products to human society. Current forgery detection models and methods rely mainly on researchers analyzing the flaws and weaknesses of forgery technologies and designing corresponding solutions. Developing self-evolving frameworks using adversarial learning mechanisms and reinforcement learning models to drive the autonomous evolution of forgery detection models, thereby improving the ability to quickly respond to various forgery products, is a key research direction for the future.
\end{itemize}
\begin{figure}[!t]
  \centering
\includegraphics[width=1.0\linewidth]{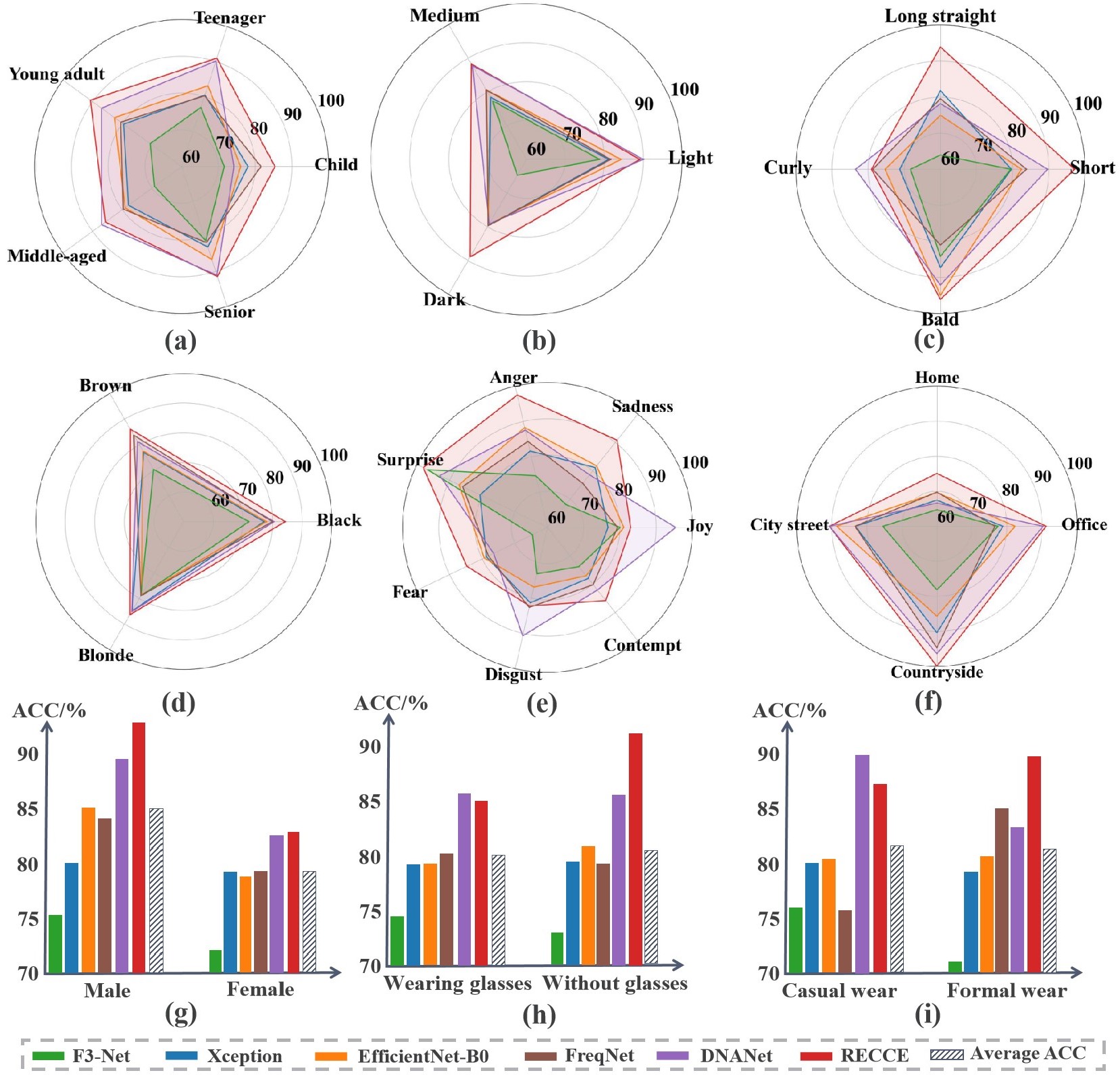} 
  \caption{Comparative evaluation of various forgery detection techniques on image-level samples from different attribute perspectives, including
 (a) age attribute, (b) skin tone attribute, (c) hair style attribute,(d) hair color attribute, (e) expression attribute, (f) background attribute, (g) gender attribute, (h) glasses attribute, and (i) dressing style attribute.
 }
  \label{figshuxing_image}
\end{figure}

\begin{figure}[!t]
  \centering
\includegraphics[width=1.0\linewidth]{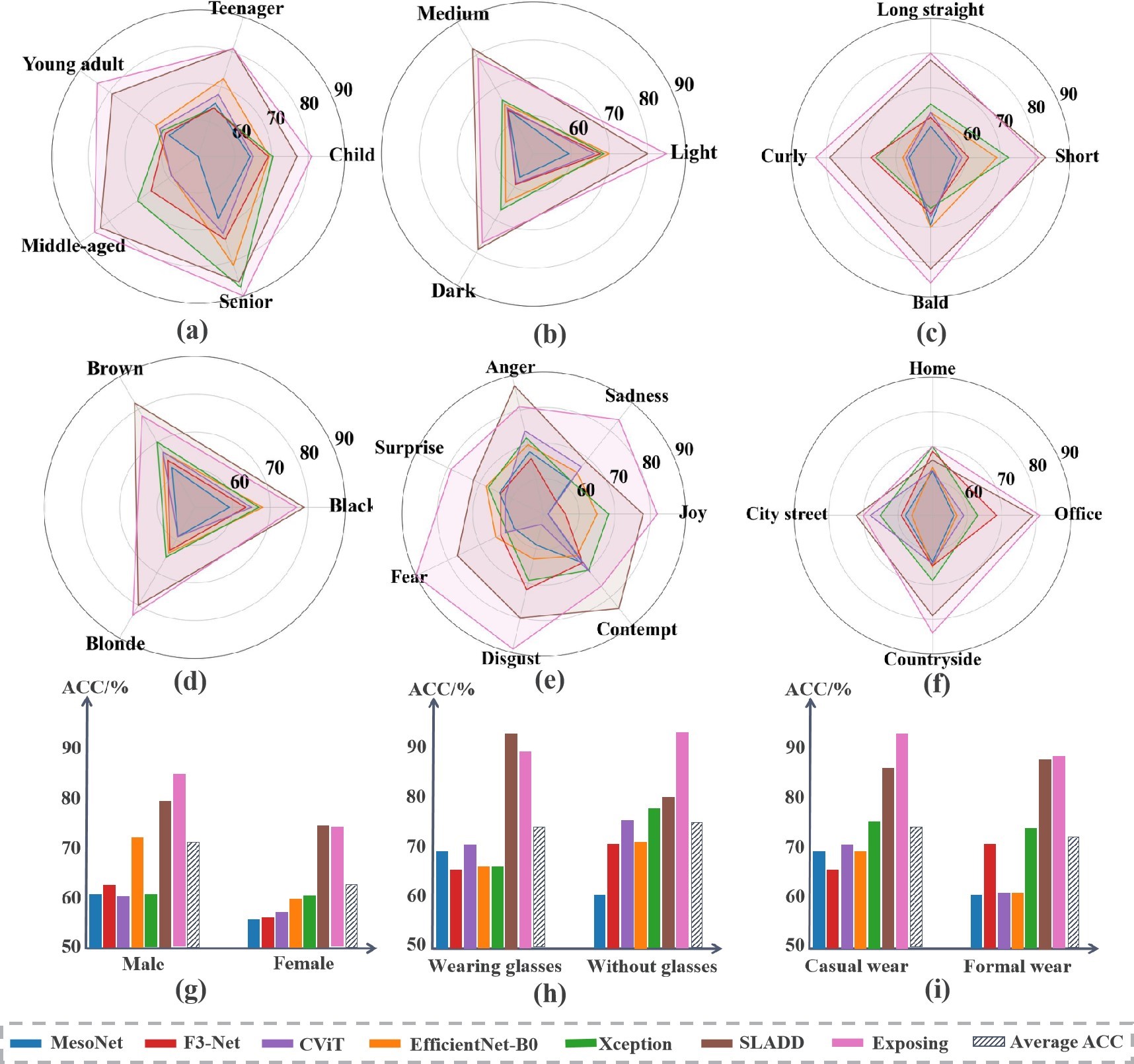} 
  \caption{Comparative evaluation of various forgery detection techniques on video-level samples from different attribute perspectives, including
 (a) age attribute, (b) skin tone attribute, (c) hair style attribute,(d) hair color attribute, (e) expression attribute, (f) background attribute, (g) gender attribute, (h) glasses attribute, and (i) dressing style attribute.
 } 
  \label{figshuxing_video}
\end{figure}
\subsection{Fine-grained Attribute Statistic Analysis for Different Forgery Techniques}
\label{surverypart8}
In this section, we train all forgery detection models using the training samples obtained from DeepFaceGen. Subsequently, we utilize the fine-grained labels provided by DeepFaceGen to conduct a detailed analysis of the detection patterns of the forgery detection techniques across $9$ attributes.

\textbf{Age Attribute.} The age attribute significantly impacts the effectiveness of forgery detection models. Figures~\ref{figshuxing_image} (a) and \ref{figshuxing_video} (a) indicate that forgery detection models face more challenges with detecting forgery samples of children, while it is easier to detect forgery data of elderly faces. This difference is due to the unique facial characteristics of children and the elderly. Children's facial features are finer and smoother, lacking prominent wrinkles and details, which makes it easier for forgery techniques to generate realistic child faces, thereby increasing the difficulty of detection. In contrast, elderly individuals often have more pronounced and complex facial features, including wrinkles, age spots, and sagging skin, which make forgery more challenging and, therefore, more likely to be detected by the model.

\textbf{Skin Tone Attribute.} The effectiveness of forgery detection models varies with different skin tones. Figures~\ref{figshuxing_image} (b) and \ref{figshuxing_video} (b) show that these models have greater difficulty in accurately detecting forgeries in individuals with darker skin tones compared to those with lighter skin tones. This highlights a racial bias inherent in the forgery detection techniques. The potential cause of this bias could be linked to variations in skin tones and the influence of lighting conditions. Individuals with darker skin tones may have facial features that are harder to capture in forgery detection. Darker skin tones can result in lower contrast in facial details, such as shadows and highlights, making it difficult for forgery detection models to identify forgery artifacts. Conversely, the facial features of individuals with lighter skin tones are generally easier to capture in images. Lighter skin tones make facial details, such as wrinkles and subtle expressions, more visible and typically maintain better facial detail contrast under various lighting conditions.

\textbf{Hair Style Attribute.} The variety of people's hairstyles also has an impact on the effectiveness of forgery detection. As shown in Figures~\ref{figshuxing_image} (c) and \ref{figshuxing_video} (c), detecting forgeries with the curly hair attribute is more difficult, while detecting those with the bald attribute is easier. In video-level experiments, the detection performance is relatively consistent across different attributes. We infer that curly hair, with its highly complex and irregular structure, contains rich details between strands. This complexity poses a greater challenge for forgery techniques in generating curly hair, making it easier to leave behind subtle artifacts that are difficult to detect. Consequently, detection models struggle to differentiate these subtle differences, increasing the difficulty of detecting forgeries with curly hair. In contrast, forgery techniques tend to produce more consistent results when generating bald heads due to the lack of complex hair structures, making it easier for detection models to identify forgery artifacts. Additionally, in video-level experiments, the continuity and motion information assist the forgery detection models in capturing forgery artifacts more effectively, leading to more balanced detection performance across different hair style attributes.

\textbf{Hair Color Attribute.} Figure~\ref{figshuxing_image} (d) and Figure~\ref{figshuxing_video} (d) show that forgery detection models perform relatively evenly when detecting forged data with the attributes of brown hair, blonde hair, and black hair. This can be attributed to similar details and contrast under lighting conditions. When generating forged images, forgery techniques typically handle similar textures and lighting effects for all three hair colors. This similarity results in detection models not having significant difficulty differences in identifying these forgeries.

\textbf{Expression Attribute.} People's inner emotions can be externalized into different expressions. Based on (e) in Figure~\ref{figshuxing_image} and Figure~\ref{figshuxing_video}, it is apparent that forgery detection models perform well when detecting forged images with the anger and surprise attributes. This may result from the facial expressions of anger and surprise attributes. They contain rich details and features that are easier to extract and recognize in image processing. Tense facial muscles and deep wrinkles are typical features of anger, while an open mouth and raised eyebrows are clear indicators of surprise. Forgery detection models can use these prominent features to enhance detection accuracy.

\textbf{Background Attribute.} The background in images/videos also influences the performance of forgery detection models. Figures~\ref{figshuxing_image} (f) and~\ref{figshuxing_video} (f) indicate that forgery detection models find it easier to detect forged images with the countryside attribute and harder to detect those with the home attribute. Background complexity may be a direct factor. Countryside backgrounds generally have lower complexity, featuring large natural landscapes such as fields, trees, and skies. These elements are relatively simple and have fewer variations, making it easier for forgery techniques to generate these backgrounds without introducing complex artifacts. Consequently, detection models can more easily identify forged elements in these simple backgrounds. By contrast, home backgrounds typically include many details and complex objects such as furniture, appliances, and decorations. Detection models need to process more details and variations, making it harder to detect forgeries.

\textbf{Gender Attribute.} The accuracy of forgery detection models is often lower for female samples ((g) in Figure~\ref{figshuxing_image} and \ref{figshuxing_video}). Similar to children in age attribute, female facial features are generally finer and smoother, lacking prominent wrinkles and rough skin texture. These fine features may make it harder for detection models to capture forgery artifacts. Additionally, women tend to wear makeup in greater numbers than men. Cosmetics can enhance or conceal certain facial features, and introduce artificial details such as eyeliner and lipstick. These changes can also make it more challenging for forgery detection models to distinguish between real and forged images, as the makeup may mask subtle forgery artifacts that the model relies on for detection.

\textbf{Glasses Attribute.} Based on Figure~\ref{figshuxing_image} (h) and Figure~\ref{figshuxing_video} (h), forgery detection models perform similarly when detecting forged data with and without the glasses attribute. This can be attributed to glasses' simple and fixed geometric features (such as frames and lenses). When generating faces with glasses, forgery techniques can maintain the stability of these geometric features well, resulting in forged images of similar quality to those without glasses.

\textbf{Dressing Style Attribute.} It can be found from Figure~\ref{figshuxing_image} (i) and Figure~\ref{figshuxing_video} (i) that forgery detection models perform similarly when detecting forged data with the casual wear attribute and the formal wear attribute. This may due to their similar complexity. Although casual and formal wear differ in style, the complexity of details in both types of clothing is relatively similar. Formal wear may include more details (such as ties and buttons), but these details do not significantly affect the quality of forged images. Casual wear may have more varied styles, but its complexity is comparable to formal wear. 

\subsection{Potential Negative Social Impacts}
\label{surverypart9}
The creation and use of deepfake datasets, while beneficial for advancing technology, can lead to several negative societal impacts:
\begin{itemize}
    \item \textbf{Misuse of Forgery Methods.} In order to restore the complex forgery scenes in the real scene as much as possible, the forgery methods in the data set are realistic. These forgery methods can be misused to create misleading or harmful content, eroding public trust in media and making it difficult to distinguish between real and fake information.
    \item  \textbf{Ethical Concerns.} Due to the transparency of the data set, a large number of face samples in the data set may provide fake resources for illegal personnel. Widespread exposure to deepfakes can lead to public skepticism and paranoia about the authenticity of all digital content.
\end{itemize}
To mitigate these impacts, we are contemplating controlled access for users and are committed to the dynamic evolution of DeepFaceGen to ensure it remains robust against emerging threats.


\begin{thebibliography}{100}

\bibitem{r1}
Yihan Cao, Siyu Li, Yixin Liu, Zhiling Yan, Yutong Dai, Philip~S. Yu, and Lichao Sun.
\newblock A comprehensive survey of ai-generated content (aigc): A history of generative ai from gan to chatgpt.
\newblock {\em ArXiv}, abs/2303.04226, 2023.

\bibitem{r2}
Wayne~Xin Zhao, Kun Zhou, Junyi Li, Tianyi Tang, Xiaolei Wang, Yupeng Hou, Yingqian Min, Beichen Zhang, Junjie Zhang, Zican Dong, Yifan Du, Chen Yang, Yushuo Chen, Zhipeng Chen, Jinhao Jiang, Ruiyang Ren, Yifan Li, Xinyu Tang, Zikang Liu, Peiyu Liu, Jian-Yun Nie, and Ji-Rong Wen.
\newblock A survey of large language models, 2023.

\bibitem{dalle2020}
{Open AI}.
\newblock {DALL·E}.
\newblock \url{https://openai.com/index/dall-e-3}, 2020.

\bibitem{sora2024}
{Open AI}.
\newblock Sora.
\newblock \url{https://openai.com/index/sora}, 2024.

\bibitem{r3}
Jonathan Ho, Ajay Jain, and Pieter Abbeel.
\newblock Denoising diffusion probabilistic models.
\newblock In H.~Larochelle, M.~Ranzato, R.~Hadsell, M.F. Balcan, and H.~Lin, editors, {\em Advances in Neural Information Processing Systems(NeurIPS)}, volume~33, pages 6840--6851. Curran Associates, Inc., 2020.

\bibitem{r10}
Falko Matern, Christian Riess, and Marc Stamminger.
\newblock Exploiting visual artifacts to expose deepfakes and face manipulations.
\newblock In {\em 2019 IEEE Winter Applications of Computer Vision Workshops (WACVW)}, pages 83--92, 2019.

\bibitem{r20}
FakeApp.
\newblock Fakeapp.
\newblock \url{https://www.deepfakescn.com}, 2019.

\bibitem{r4}
Apurva Gandhi and Shomik Jain.
\newblock Adversarial perturbations fool deepfake detectors.
\newblock In {\em 2020 International Joint Conference on Neural Networks (IJCNN)}, pages 1--8, 2020.

\bibitem{r13}
Yuezun Li, Xin Yang, Pu~Sun, Honggang Qi, and Siwei Lyu.
\newblock Celeb-df: A large-scale challenging dataset for deepfake forensics.
\newblock In {\em Proceedings of the IEEE/CVF Conference on Computer Vision and Pattern Recognition (CVPR)}, June 2020.

\bibitem{r14}
Liming Jiang, Ren Li, Wayne Wu, Chen Qian, and Chen~Change Loy.
\newblock Deeperforensics-1.0: A large-scale dataset for real-world face forgery detection.
\newblock In {\em Proceedings of the IEEE/CVF Conference on Computer Vision and Pattern Recognition (CVPR)}, June 2020.

\bibitem{r11}
Pavel Korshunov and S{\'e}bastien Marcel.
\newblock Deepfakes: a new threat to face recognition? assessment and detection.
\newblock {\em ArXiv}, abs/1812.08685, 2018.

\bibitem{r12}
Andreas Rossler, Davide Cozzolino, Luisa Verdoliva, Christian Riess, Justus Thies, and Matthias Niessner.
\newblock Faceforensics++: Learning to detect manipulated facial images.
\newblock In {\em Proceedings of the IEEE/CVF International Conference on Computer Vision (ICCV)}, October 2019.

\bibitem{r22}
{Faceswap}.
\newblock Faceswap: Deepfakes software for all.
\newblock \url{https://github.com/deepfakes/faceswap}, 2020.

\bibitem{r23}
Justus Thies, Michael Zollhöfer, Marc Stamminger, Christian Theobalt, and Matthias Nießner.
\newblock Face2face: Real-time face capture and reenactment of rgb videos.
\newblock In {\em 2016 IEEE Conference on Computer Vision and Pattern Recognition (CVPR)}, pages 2387--2395, 2016.

\bibitem{r24}
{Faceswap}.
\newblock Faceswap.
\newblock \url{https://github.com/MarekKowalski/FaceSwap/}, 2020.

\bibitem{r25}
Justus Thies, Michael Zollhfer, and Matthias Niener.
\newblock Deferred neural rendering: Image synthesis using neural textures.
\newblock {\em ACM Transactions on Graphics (ACM Trans. Graph)}, 2019.

\bibitem{r15}
Bojia Zi, Minghao Chang, Jingjing Chen, Xingjun Ma, and Yu-Gang Jiang.
\newblock Wilddeepfake: A challenging real-world dataset for deepfake detection.
\newblock {\em Proceedings of the 28th ACM International Conference on Multimedia}, 2020.

\bibitem{r16}
Brian Dolhansky, Joanna Bitton, Ben Pflaum, Jikuo Lu, Russ Howes, Menglin Wang, and Cristian~Canton Ferrer.
\newblock The deepfake detection challenge (dfdc) dataset, 2020.

\bibitem{r17}
Patrick Kwon, Jaeseong You, Gyuhyeon Nam, Sungwoo Park, and Gyeongsu Chae.
\newblock Kodf: A large-scale korean deepfake detection dataset.
\newblock {\em 2021 IEEE/CVF International Conference on Computer Vision (ICCV)}, pages 10724--10733, 2021.

\bibitem{r18}
Hasam Khalid, Shahroz Tariq, Minha Kim, and Simon~S. Woo.
\newblock Fakeavceleb: A novel audio-video multimodal deepfake dataset, 2022.

\bibitem{r26}
Ivan Petrov, Daiheng Gao, Nikolay Chervoniy, Kunlin Liu, Sugasa Marangonda, Chris Um{\'e}, Mr. Dpfks, RP~Luis, Jian Jiang, Sheng Zhang, Pingyu Wu, Bo~Zhou, and Weiming Zhang.
\newblock Deepfacelab: A simple, flexible and extensible face swapping framework.
\newblock {\em ArXiv}, abs/2005.05535, 2020.

\bibitem{r27}
Yuval Nirkin, Yosi Keller, and Tal Hassner.
\newblock Fsgan: Subject agnostic face swapping and reenactment.
\newblock In {\em 2019 IEEE/CVF International Conference on Computer Vision (ICCV)}, pages 7183--7192, 2019.

\bibitem{r19}
Yinan He, Bei Gan, Siyu Chen, Yichun Zhou, Guojun Yin, Luchuan Song, Lu~Sheng, Jing Shao, and Ziwei Liu.
\newblock Forgerynet: A versatile benchmark for comprehensive forgery analysis.
\newblock In {\em 2021 IEEE/CVF Conference on Computer Vision and Pattern Recognition (CVPR)}, pages 4358--4367, 2021.

\bibitem{midjourney2022}
{Midjourney}.
\newblock Midjourney.
\newblock \url{https://www.midjourney.com/home}, 2022.

\bibitem{r6}
Davide Cozzolino, Koki Nagano, Lucas Thomaz, Angshul Majumdar, and Luisa Verdoliva.
\newblock Synthetic image detection: Highlights from the ieee video and image processing cup 2022 student competition.
\newblock {\em IEEE Signal Processing Magazine}, 40(7):94--100, 2023.

\bibitem{r7}
Zeyang Sha, Zheng Li, Ning Yu, and Yang Zhang.
\newblock De-fake: Detection and attribution of fake images generated by text-to-image generation models.
\newblock In {\em Proceedings of the 2023 ACM SIGSAC Conference on Computer and Communications Security}, CCS '23, page 3418–3432, New York, NY, USA, 2023. Association for Computing Machinery.

\bibitem{2009Learning}
A.~Krizhevsky and G.~Hinton.
\newblock Learning multiple layers of features from tiny images.
\newblock {\em Handbook of Systemic Autoimmune Diseases}, 1(4), 2009.

\bibitem{r8}
Jordan~J. Bird and Ahmad Lotfi.
\newblock Cifake: Image classification and explainable identification of ai-generated synthetic images.
\newblock {\em IEEE Access}, 12:15642--15650, 2024.

\bibitem{r9}
Mingjian Zhu, Hanting Chen, Qiangyu Yan, Xudong Huang, Guanyu Lin, Wei Li, Zhijun Tu, Hailin Hu, Jie Hu, and Yunhe Wang.
\newblock Genimage: A million-scale benchmark for detecting ai-generated image, 2023.

\bibitem{brock2018large}
Andrew Brock, Jeff Donahue, and Karen Simonyan.
\newblock Large scale {GAN} training for high fidelity natural image synthesis.
\newblock In {\em International Conference on Learning Representations (ICLR)}, 2019.

\bibitem{r5}
Yabin Wang, Zhiwu Huang, and Xiaopeng Hong.
\newblock Benchmarking deepart detection.
\newblock {\em ArXiv}, abs/2302.14475, 2023.

\bibitem{DFGAN}
Ming Tao, Hao Tang, Fei Wu, Xiaoyuan Jing, Bing-Kun Bao, and Changsheng Xu.
\newblock Df-gan: A simple and effective baseline for text-to-image synthesis.
\newblock In {\em 2022 IEEE/CVF Conference on Computer Vision and Pattern Recognition (CVPR)}, pages 16494--16504, 2022.

\bibitem{wenxin2022}
{Baidu}.
\newblock Wenxin.
\newblock \url{https://yige.baidu.com/}, 2022.

\bibitem{SD}
{Stability.ai}.
\newblock Stable {D}iffusion.
\newblock \url{https://stability.ai/}, 2023.

\bibitem{openjourney}
{PromptHero}.
\newblock Openjourney.
\newblock \url{http://openjourney.art/}, 2023.

\bibitem{hu2022lora}
Edward~J Hu, Yelong Shen, Phillip Wallis, Zeyuan Allen-Zhu, Yuanzhi Li, Shean Wang, Lu~Wang, and Weizhu Chen.
\newblock Lo{RA}: Low-rank adaptation of large language models.
\newblock In {\em International Conference on Learning Representations (ICLR)}, 2022.

\bibitem{yuan2024magictime}
Shenghai Yuan, Jinfa Huang, Yujun Shi, Yongqi Xu, Ruijie Zhu, Bin Lin, Xinhua Cheng, Li~Yuan, and Jiebo Luo.
\newblock Magictime: Time-lapse video generation models as metamorphic simulators.
\newblock {\em arXiv preprint arXiv:2404.05014}, 2024.

\bibitem{ACMlightning}
Shanchuan Lin and Xiao Yang.
\newblock Animatediff-lightning: Cross-model diffusion distillation, 2024.

\bibitem{wang2024animatelcm}
Fu-Yun Wang, Zhaoyang Huang, Xiaoyu Shi, Weikang Bian, Guanglu Song, Yu~Liu, and Hongsheng Li.
\newblock Animatelcm: Accelerating the animation of personalized diffusion models and adapters with decoupled consistency learning, 2024.

\bibitem{Mullan_Hotshot-XL_2023}
John Mullan, Duncan Crawbuck, and Aakash Sastry.
\newblock {Hotshot-XL}.
\newblock \url{https://github.com/hotshotco/hotshot-xl}, 2023.

\bibitem{zeroscope_v2_576w}
{Academy for Discovery, Adventure, Momentum and Outlook}.
\newblock Zeroscope.
\newblock \url{https://huggingface.co/cerspense/zeroscope_v2_576w}, 2023.

\bibitem{Li2019FaceShifterTH}
Lingzhi Li, Jianmin Bao, Hao Yang, Dong Chen, and Fang Wen.
\newblock Faceshifter: Towards high fidelity and occlusion aware face swapping.
\newblock {\em ArXiv}, abs/1912.13457, 2019.

\bibitem{blendface}
Kaede Shiohara, Xingchao Yang, and Takafumi Taketomi.
\newblock Blendface: Re-designing identity encoders for face-swapping.
\newblock In {\em Proceedings of the IEEE/CVF International Conference on Computer Vision (ICCV)}, pages 7634--7644, October 2023.

\bibitem{Chen2020SimSwapAE}
Renwang Chen, Xuanhong Chen, Bingbing Ni, and Yanhao Ge.
\newblock Simswap: An efficient framework for high fidelity face swapping.
\newblock {\em Proceedings of the 28th ACM International Conference on Multimedia}, 2020.

\bibitem{talkinghead}
Ohad Fried, Ayush Tewari, Michael Zollh\"{o}fer, Adam Finkelstein, Eli Shechtman, Dan~B Goldman, Kyle Genova, Zeyu Jin, Christian Theobalt, and Maneesh Agrawala.
\newblock Text-based editing of talking-head video.
\newblock {\em ACM Trans. Graph.}, 38(4), jul 2019.

\bibitem{atvgnet}
Lele Chen, Ross~K Maddox, Zhiyao Duan, and Chenliang Xu.
\newblock Hierarchical cross-modal talking face generation with dynamic pixel-wise loss.
\newblock In {\em Proceedings of the IEEE Conference on Computer Vision and Pattern Recognition (CVPR)}, pages 7832--7841, 2019.

\bibitem{FOMM}
Aliaksandr Siarohin, Stéphane Lathuilière, Sergey Tulyakov, Elisa Ricci, and Nicu Sebe.
\newblock First order motion model for image animation.
\newblock In {\em Conference on Neural Information Processing Systems (NeurIPS)}, December 2019.

\bibitem{motion}
Aliaksandr Siarohin, Subhankar Roy, Stéphane Lathuilière, Sergey Tulyakov, Elisa Ricci, and Nicu Sebe.
\newblock Motion supervised co-part segmentation.
\newblock {\em arXiv preprint}, 2020.

\bibitem{StyleGAN2}
Tero Karras, Samuli Laine, Miika Aittala, Janne Hellsten, Jaakko Lehtinen, and Timo Aila.
\newblock Analyzing and improving the image quality of stylegan.
\newblock {\em 2020 IEEE/CVF Conference on Computer Vision and Pattern Recognition (CVPR)}, pages 8107--8116, 2019.

\bibitem{maskgan}
Cheng-Han Lee, Ziwei Liu, Lingyun Wu, and Ping Luo.
\newblock Maskgan: Towards diverse and interactive facial image manipulation.
\newblock {\em 2020 IEEE/CVF Conference on Computer Vision and Pattern Recognition (CVPR)}, pages 5548--5557, 2019.

\bibitem{StarGAN2}
Yunjey Choi, Youngjung Uh, Jaejun Yoo, and Jung-Woo Ha.
\newblock Stargan v2: Diverse image synthesis for multiple domains.
\newblock {\em 2020 IEEE/CVF Conference on Computer Vision and Pattern Recognition (CVPR)}, pages 8185--8194, 2019.

\bibitem{scfegan}
Youngjoo Jo and Jongyoul Park.
\newblock Sc-fegan: Face editing generative adversarial network with user's sketch and color.
\newblock In {\em The IEEE International Conference on Computer Vision (ICCV)}, October 2019.

\bibitem{DiscoFaceGAN}
Yu~Deng, Jiaolong Yang, Dong Chen, Fang Wen, and Xin Tong.
\newblock Disentangled and controllable face image generation via 3d imitative-contrastive learning.
\newblock In {\em 2020 IEEE/CVF Conference on Computer Vision and Pattern Recognition (CVPR)}, pages 5153--5162, 2020.

\bibitem{CN-CVS}
Chen Chen, Dong Wang, and Thomas~Fang Zheng.
\newblock Cn-cvs: A mandarin audio-visual dataset for large vocabulary continuous visual to speech synthesis.
\newblock In {\em ICASSP 2023 - 2023 IEEE International Conference on Acoustics, Speech and Signal Processing (ICASSP)}, pages 1--5, 2023.

\bibitem{CMLR}
Ya~Zhao, Rui Xu, and Mingli Song.
\newblock A cascade sequence-to-sequence model for chinese mandarin lip reading.
\newblock {\em ACM}, 2019.

\bibitem{skin}
René Alejandro~Rejón Pia and Chenglong Ma.
\newblock Classification algorithm for skin color (casco): A new tool to measure skin color in social science research.
\newblock {\em Social Science Quarterly}, 104:168, 2023.

\bibitem{yolo}
{ultralytics}.
\newblock Yolov5.
\newblock \url{https://github.com/ultralytics/yolov5}, 2020.

\bibitem{Xception}
François Chollet.
\newblock Xception: Deep learning with depthwise separable convolutions.
\newblock In {\em 2017 IEEE Conference on Computer Vision and Pattern Recognition (CVPR)}, pages 1800--1807, 2017.

\bibitem{tan2020efficientnet}
Mingxing Tan and Quoc~V. Le.
\newblock Efficientnet: Rethinking model scaling for convolutional neural networks, 2020.

\bibitem{F3NET}
Yuyang Qian, Guojun Yin, Lu~Sheng, Zixuan Chen, and Jing Shao.
\newblock Thinking in frequency: Face forgery detection by mining frequency-aware clues, 2020.

\bibitem{recce}
Junyi Cao, Chao Ma, Taiping Yao, Shen Chen, Shouhong Ding, and Xiaokang Yang.
\newblock End-to-end reconstruction-classification learning for face forgery detection.
\newblock In {\em Proceedings of the IEEE/CVF Conference on Computer Vision and Pattern Recognition (CVPR)}, pages 4113--4122, June 2022.

\bibitem{dnaet}
Tianyun Yang, Ziyao Huang, Juan Cao, Lei Li, and Xirong Li.
\newblock Deepfake network architecture attribution.
\newblock In {\em Proceedings of the 36th AAAI Conference on Artificial Intelligence (AAAI)}, 2022.

\bibitem{tan2024frequencyaware}
Chuangchuang Tan, Yao Zhao, Shikui Wei, Guanghua Gu, Ping Liu, and Yunchao Wei.
\newblock Frequency-aware deepfake detection: Improving generalizability through frequency space learning, 2024.

\bibitem{Afchar2018MesoNetAC}
Darius Afchar, Vincent Nozick, Junichi Yamagishi, and Isao Echizen.
\newblock Mesonet: a compact facial video forgery detection network.
\newblock {\em 2018 IEEE International Workshop on Information Forensics and Security (WIFS)}, pages 1--7, 2018.

\bibitem{cvit}
Deressa Wodajo and Solomon Atnafu.
\newblock Deepfake video detection using convolutional vision transformer, 2021.

\bibitem{sladd}
Liang Chen, Yong Zhang, Yibing Song, Lingqiao Liu, and Jue Wang.
\newblock Self-supervised learning of adversarial examples: Towards good generalizations for deepfake detections.
\newblock In {\em CVPR}, 2022.

\bibitem{ba2024exposing}
Zhongjie Ba, Qingyu Liu, Zhenguang Liu, Shuang Wu, Feng Lin, Li~Lu, and Kui Ren.
\newblock Exposing the deception: Uncovering more forgery clues for deepfake detection, 2024.

\bibitem{TGAN}
Masaki Saito, Eiichi Matsumoto, and Shunta Saito.
\newblock Temporal generative adversarial nets with singular value clipping.
\newblock In {\em 2017 IEEE International Conference on Computer Vision (ICCV)}, pages 2849--2858, 2017.

\bibitem{DVD-GAN}
Aidan Clark, Jeff Donahue, and Karen Simonyan.
\newblock Efficient video generation on complex datasets.
\newblock {\em ArXiv}, abs/1907.06571, 2019.

\bibitem{videogpt}
Wilson Yan, Yunzhi Zhang, Pieter Abbeel, and Aravind Srinivas.
\newblock Videogpt: Video generation using vq-vae and transformers, 2021.

\bibitem{t-SNE}
Laurens van~der Maaten and Geoffrey~E. Hinton.
\newblock Visualizing data using t-sne.
\newblock {\em Journal of Machine Learning Research}, 9:2579--2605, 2008.

\bibitem{resnet}
Kaiming He, Xiangyu Zhang, Shaoqing Ren, and Jian Sun.
\newblock Deep residual learning for image recognition.
\newblock In {\em 2016 IEEE Conference on Computer Vision and Pattern Recognition (CVPR)}, pages 770--778, 2016.

\bibitem{shaoanlu2017}
Shaoanlu.
\newblock Faceswap-gan.
\newblock \url{https://github.com/shaoanlu/faceswap-GAN}, 2017.
\newblock CP/OL, accessed 2021-10-15.

\bibitem{Natsume2018RSGANFS}
Ryota Natsume, Tatsuya Yatagawa, and Shigeo Morishima.
\newblock Rsgan: face swapping and editing using face and hair representation in latent spaces.
\newblock {\em ACM SIGGRAPH 2018 Posters}, 2018.

\bibitem{Imaginator}
Yaohui Wang, Piotr Bilinski, Francois Bremond, and Antitza Dantcheva.
\newblock Imaginator: Conditional spatio-temporal gan for video generation.
\newblock In {\em 2020 IEEE Winter Conference on Applications of Computer Vision (WACV)}, pages 1149--1158, 2020.

\bibitem{Monkey-Net}
Aliaksandr Siarohin, Stéphane Lathuilière, Sergey Tulyakov, Elisa Ricci, and Nicu Sebe.
\newblock Animating arbitrary objects via deep motion transfer.
\newblock In {\em 2019 IEEE/CVF Conference on Computer Vision and Pattern Recognition (CVPR)}, pages 2372--2381, 2019.

\bibitem{GANimation}
Albert Pumarola, Antonio Agudo, Aleix~M. Martinez, Alberto Sanfeliu, and Francesc Moreno-Noguer.
\newblock Ganimation: Anatomically-aware facial animation from a single image.
\newblock In Vittorio Ferrari, Martial Hebert, Cristian Sminchisescu, and Yair Weiss, editors, {\em Computer Vision -- ECCV 2018}, pages 835--851, Cham, 2018. Springer International Publishing.

\bibitem{FACEGAN}
Soumya Tripathy, Juho Kannala, and Esa Rahtu.
\newblock Facegan: Facial attribute controllable reenactment gan, 2020.

\bibitem{CycleGAN}
Jun-Yan Zhu, Taesung Park, Phillip Isola, and Alexei~A. Efros.
\newblock Unpaired image-to-image translation using cycle-consistent adversarial networks.
\newblock In {\em 2017 IEEE International Conference on Computer Vision (ICCV)}, pages 2242--2251, 2017.

\bibitem{Xu2017FaceTW}
Runze Xu, Zhiming Zhou, Weinan Zhang, and Yong Yu.
\newblock Face transfer with generative adversarial network.
\newblock {\em ArXiv}, abs/1710.06090, 2017.

\bibitem{Recycle-GAN}
Aayush Bansal, Shugao Ma, Deva Ramanan, and Yaser Sheikh.
\newblock Recycle-gan: Unsupervised video retargeting.
\newblock In {\em ECCV}, 2018.

\bibitem{ReenactGAN}
Wayne Wu, Yunxuan Zhang, Cheng Li, Chen Qian, and Chen~Change Loy.
\newblock Reenactgan: Learning to reenact faces via boundary transfer.
\newblock In {\em ECCV}, 2018.

\bibitem{StyleGAN1}
Tero Karras, Samuli Laine, and Timo Aila.
\newblock A style-based generator architecture for generative adversarial networks.
\newblock {\em 2019 IEEE/CVF Conference on Computer Vision and Pattern Recognition (CVPR)}, pages 4396--4405, 2018.

\bibitem{StyleGAN3}
Tero Karras, Miika Aittala, Samuli Laine, Erik H\"{a}rk\"{o}nen, Janne Hellsten, Jaakko Lehtinen, and Timo Aila.
\newblock Alias-free generative adversarial networks.
\newblock In M.~Ranzato, A.~Beygelzimer, Y.~Dauphin, P.S. Liang, and J.~Wortman Vaughan, editors, {\em Advances in Neural Information Processing Systems}, volume~34, pages 852--863. Curran Associates, Inc., 2021.

\bibitem{StarGAN1}
Yunjey Choi, Min-Je Choi, Mun~Su Kim, Jung-Woo Ha, Sunghun Kim, and Jaegul Choo.
\newblock Stargan: Unified generative adversarial networks for multi-domain image-to-image translation.
\newblock {\em 2018 IEEE/CVF Conference on Computer Vision and Pattern Recognition(CVPR)}, pages 8789--8797, 2017.

\bibitem{GANnotation}
Enrique Sanchez and Michel~F. Valstar.
\newblock Triple consistency loss for pairing distributions in gan-based face synthesis.
\newblock {\em ArXiv}, abs/1811.03492, 2018.

\bibitem{CAM}
Daejin Kim, Mohammad~Azam Khan, and Jaegul Choo.
\newblock Not just compete, but collaborate: Local image-to-image translation via cooperative mask prediction.
\newblock In {\em 2021 IEEE/CVF Conference on Computer Vision and Pattern Recognition (CVPR)}, pages 6505--6514, 2021.

\bibitem{HiSD}
Xinyang Li, Shengchuan Zhang, Jie Hu, Liujuan Cao, Xiaopeng Hong, Xudong Mao, Feiyue Huang, Yongjian Wu, and Rongrong Ji.
\newblock Image-to-image translation via hierarchical style disentanglement.
\newblock In {\em 2021 IEEE/CVF Conference on Computer Vision and Pattern Recognition (CVPR)}, pages 8635--8644, 2021.

\bibitem{GOGAN}
Musadaq Mansoor, Mohammad Nauman, Hafeez~Ur Rehman, and Alfredo Benso.
\newblock Gene ontology gan (gogan): a novel architecture for protein function prediction.
\newblock {\em Soft Computing}, 26(16):7653--7667, August 2022.

\bibitem{ding2021cogview}
Ming Ding, Zhuoyi Yang, Wenyi Hong, Wendi Zheng, Chang Zhou, Da~Yin, Junyang Lin, Xu~Zou, Zhou Shao, Hongxia Yang, and Jie Tang.
\newblock Cogview: Mastering text-to-image generation via transformers.
\newblock {\em arXiv preprint arXiv:2105.13290}, 2021.

\bibitem{Make-A-Scene}
Oran Gafni, Adam Polyak, Oron Ashual, Shelly Sheynin, Devi Parikh, and Yaniv Taigman.
\newblock Make-a-scene: Scene-based text-to-image generation with human priors, 2022.

\bibitem{GLIDE}
Alexander~Quinn Nichol, Prafulla Dhariwal, Aditya Ramesh, Pranav Shyam, Pamela Mishkin, Bob Mcgrew, Ilya Sutskever, and Mark Chen.
\newblock {GLIDE}: Towards photorealistic image generation and editing with text-guided diffusion models.
\newblock In Kamalika Chaudhuri, Stefanie Jegelka, Le~Song, Csaba Szepesvari, Gang Niu, and Sivan Sabato, editors, {\em Proceedings of the 39th International Conference on Machine Learning}, volume 162 of {\em Proceedings of Machine Learning Research}, pages 16784--16804. PMLR, 17--23 Jul 2022.

\bibitem{Imagen}
Chitwan Saharia, William Chan, Saurabh Saxena, Lala Li, Jay Whang, Emily Denton, Seyed Kamyar~Seyed Ghasemipour, Burcu~Karagol Ayan, S.~Sara Mahdavi, Rapha~Gontijo Lopes, Tim Salimans, Jonathan Ho, David~J Fleet, and Mohammad Norouzi.
\newblock Photorealistic text-to-image diffusion models with deep language understanding, 2022.

\bibitem{stable}
Robin Rombach, Andreas Blattmann, Dominik Lorenz, Patrick Esser, and Björn Ommer.
\newblock High-resolution image synthesis with latent diffusion models, 2021.

\bibitem{54}
Hanqing Zhao, Tianyi Wei, Wenbo Zhou, Weiming Zhang, Dongdong Chen, and Nenghai Yu.
\newblock Multi-attentional deepfake detection.
\newblock In {\em 2021 IEEE/CVF Conference on Computer Vision and Pattern Recognition (CVPR)}, pages 2185--2194, 2021.

\bibitem{55}
Lingzhi Li, Jianmin Bao, Ting Zhang, Hao Yang, Dong Chen, Fang Wen, and Baining Guo.
\newblock Face x-ray for more general face forgery detection.
\newblock In {\em 2020 IEEE/CVF Conference on Computer Vision and Pattern Recognition (CVPR)}, pages 5000--5009, 2020.

\bibitem{56}
Kaede Shiohara and Toshihiko Yamasaki.
\newblock Detecting deepfakes with self-blended images.
\newblock In {\em Proceedings of the IEEE/CVF Conference on Computer Vision and Pattern Recognition}, pages 18720--18729, 2022.

\bibitem{57}
Junyi Cao, Chao Ma, Taiping Yao, Shen Chen, Shouhong Ding, and Xiaokang Yang.
\newblock End-to-end reconstruction-classification learning for face forgery detection.
\newblock In {\em 2022 IEEE/CVF Conference on Computer Vision and Pattern Recognition (CVPR)}, pages 4103--4112, 2022.

\bibitem{58}
S.~Dong, Jin Wang, Jiajun Liang, Haoqiang Fan, and Renhe Ji.
\newblock Explaining deepfake detection by analysing image matching.
\newblock In {\em European Conference on Computer Vision}, 2022.

\bibitem{59}
Yuyang Qian, Guojun Yin, Lu~Sheng, Zixuan Chen, and Jing Shao.
\newblock Thinking in frequency: Face forgery detection by mining frequency-aware clues.
\newblock In {\em Computer Vision – ECCV 2020: 16th European Conference, Glasgow, UK, August 23–28, 2020, Proceedings, Part XII}, page 86–103, Berlin, Heidelberg, 2020. Springer-Verlag.

\bibitem{60}
Junke Wang, Zuxuan Wu, Wenhao Ouyang, Xintong Han, Jingjing Chen, Yu-Gang Jiang, and Ser-Nam Li.
\newblock M2tr: Multi-modal multi-scale transformers for deepfake detection.
\newblock In {\em Proceedings of the 2022 International Conference on Multimedia Retrieval}, ICMR '22, page 615–623, New York, NY, USA, 2022. Association for Computing Machinery.

\bibitem{61}
Iacopo Masi, Aditya Killekar, Royston~Marian Mascarenhas, Shenoy~Pratik Gurudatt, and Wael AbdAlmageed.
\newblock Two-branch recurrent network for isolating deepfakes in videos.
\newblock {\em ArXiv}, abs/2008.03412, 2020.

\bibitem{62}
Lukas Ruff, Nico G{\"o}rnitz, Lucas Deecke, Shoaib~Ahmed Siddiqui, Robert~A. Vandermeulen, Alexander Binder, Emmanuel M{\"u}ller, and M.~Kloft.
\newblock Deep one-class classification.
\newblock In {\em International Conference on Machine Learning}, 2018.

\bibitem{63}
Yinglin Zheng, Jianmin Bao, Dong Chen, Ming Zeng, and Fang Wen.
\newblock Exploring temporal coherence for more general video face forgery detection.
\newblock {\em 2021 IEEE/CVF International Conference on Computer Vision (ICCV)}, pages 15024--15034, 2021.

\bibitem{64}
Riccardo Corvi, Davide Cozzolino, Giada Zingarini, Giovanni Poggi, Koki Nagano, and Luisa Verdoliva.
\newblock On the detection of synthetic images generated by diffusion models.
\newblock In {\em ICASSP 2023 - 2023 IEEE International Conference on Acoustics, Speech and Signal Processing (ICASSP)}, pages 1--5, 2023.

\bibitem{DIRE}
Zhendong Wang, Jianmin Bao, Wen gang Zhou, Weilun Wang, Hezhen Hu, Hong Chen, and Houqiang Li.
\newblock Dire for diffusion-generated image detection.
\newblock {\em 2023 IEEE/CVF International Conference on Computer Vision (ICCV)}, pages 22388--22398, 2023.

\bibitem{MBConv}
Mark Sandler, Andrew Howard, Menglong Zhu, Andrey Zhmoginov, and Liang-Chieh Chen.
\newblock Mobilenetv2: Inverted residuals and linear bottlenecks.
\newblock In {\em 2018 IEEE/CVF Conference on Computer Vision and Pattern Recognition (CVPR)}, pages 4510--4520, 2018.

\bibitem{squeeze}
Jie Hu, Li~Shen, and Gang Sun.
\newblock Squeeze-and-excitation networks.
\newblock In {\em 2018 IEEE/CVF Conference on Computer Vision and Pattern Recognition (CVPR)}, pages 7132--7141, 2018.

\end{thebibliography}
\end{document}